\documentclass{article} 
\usepackage{iclr2023_conference,times}

\usepackage{amsmath,amsfonts,bm}

\def\eqref#1{equation~\ref{#1}}

\def\1{\bm{1}}

\DeclareMathAlphabet{\mathsfit}{\encodingdefault}{\sfdefault}{m}{sl}
\SetMathAlphabet{\mathsfit}{bold}{\encodingdefault}{\sfdefault}{bx}{n}

\def\gE{{\mathcal{E}}}

\def\gG{{\mathcal{G}}}

\def\gV{{\mathcal{V}}}

\usepackage{hyperref}
\usepackage{url}
\usepackage{booktabs}       
\usepackage{amsfonts}       
\usepackage{graphicx}       
\usepackage{epstopdf}       
\usepackage{multirow}       
\usepackage{subcaption}     
\usepackage{xspace} 
\usepackage{amsmath,amssymb,amsfonts} 
\usepackage{algorithm}
\usepackage{algorithmic}
\usepackage{amsthm}
\usepackage{enumitem}

\newcommand{\gnn}{\footnotesize{\textsc{GNN}}}
\newcommand{\prop}{DGE\xspace}
\newcommand{\propa}{DGE-concat}
\newcommand{\propb}{DGE-pool}
\newcommand{\propc}{DGE-bag}
\newcommand{\propd}{DGE-batch}
\newcommand{\myg}{\gG}
\newcommand{\myv}{\gV}
\newcommand{\mye}{\gE}
\newcommand{\sampv}{V}

\title{Deep Ensembles for Graphs with Higher-order Dependencies}

\author{%
  Steven~J.~Krieg, William~C.~Burgis, Patrick~M.~Soga, \& Nitesh~V.~Chawla\thanks{Corresponding author.} \\
  Lucy Family Institute for Data and Society\\
  University of Notre Dame\\
  Notre Dame, IN 46556 \\
  \texttt{\{skrieg,wburgis,psoga,nchawla\}@nd.edu}
}

\iclrfinalcopy 
\begin{document}

\maketitle

\begin{abstract}
Graph neural networks (GNNs) continue to achieve state-of-the-art performance on many graph learning tasks, but rely on the assumption that a given graph is a sufficient approximation of the true neighborhood structure. When a system contains higher-order sequential dependencies, we show that the tendency of traditional graph representations to underfit each node's neighborhood causes existing GNNs to generalize poorly. To address this, we propose a novel \textbf{Deep Graph Ensemble} (\prop), which captures neighborhood variance by training an ensemble of GNNs on different neighborhood subspaces of the same node within a higher-order network representation. We show that \prop consistently outperforms existing GNNs on semisupervised and supervised tasks on six real-world data sets with known higher-order dependencies, even under a similar parameter budget. We demonstrate that diverse and accurate base classifiers are central to \prop's success, and discuss the implications of these findings for future work on ensembles of GNNs.
\end{abstract}
\section{Introduction} \label{sec:introduction}

Graph neural networks (GNNs) solve learning tasks by propagating information through each node's neighborhood in a graph \citep{zhou2020graph,wu2020comprehensive}. Most present work on GNNs assumes that a given graph is a sufficient approximation of the underlying neighborhood structure. But a growing body of work has challenged this assumption by showing that traditional graphs often cannot capture the higher-order structure and dynamics that govern many real-world systems \citep{lambiotte2019networks,battiston2020networks,porter2020nonlinearity,torres2021and,battiston2021physics}. In the present work, we couple GNNs with a specific family of graphs, \textbf{higher-order networks} (HONs), which encode sequential \textbf{higher-order dependencies} (i.e., conditional probabilities that cannot be explained by a first-order Markov model) in a graph structure. A traditional graph, which we call a \textbf{first-order network} (FON), represents a system by decomposing it into a set of pairwise edges, so the only way to infer polyadic interactions is via transitive paths over adjacent nodes. When higher-order dependencies are present, these Markovian paths underfit the true neighborhood \citep{scholtes2017network} and can thus produce many false positive interactions between nodes \citep{lambiotte2019networks}. To address this limitation, \citet{xu2016representing} proposed a HON that creates conditional nodes to more accurately encode the observed higher-order interactions. By preserving this additional information in the graph structure, HONs have produced new insights in studies of user behavior \citep{chierichetti2012web}, citation networks \citep{rosvall2014memory}, human mobility and navigation patterns \citep{scholtes2014causality,peixoto2017modelling}, the spread of invasive species \cite{saebi2020network}, anomaly detection \citep{saebi2020efficient}, disease progression \citep{krieg2020higher}, and more \citep{koher2016infections,peixoto2017modelling,scholtes2017network,lambiotte2019networks,saebi2020honem}. However, their use with GNNs has not been thoroughly explored.

\begin{figure}[t!]
    \centering
    \includegraphics[width=\textwidth]{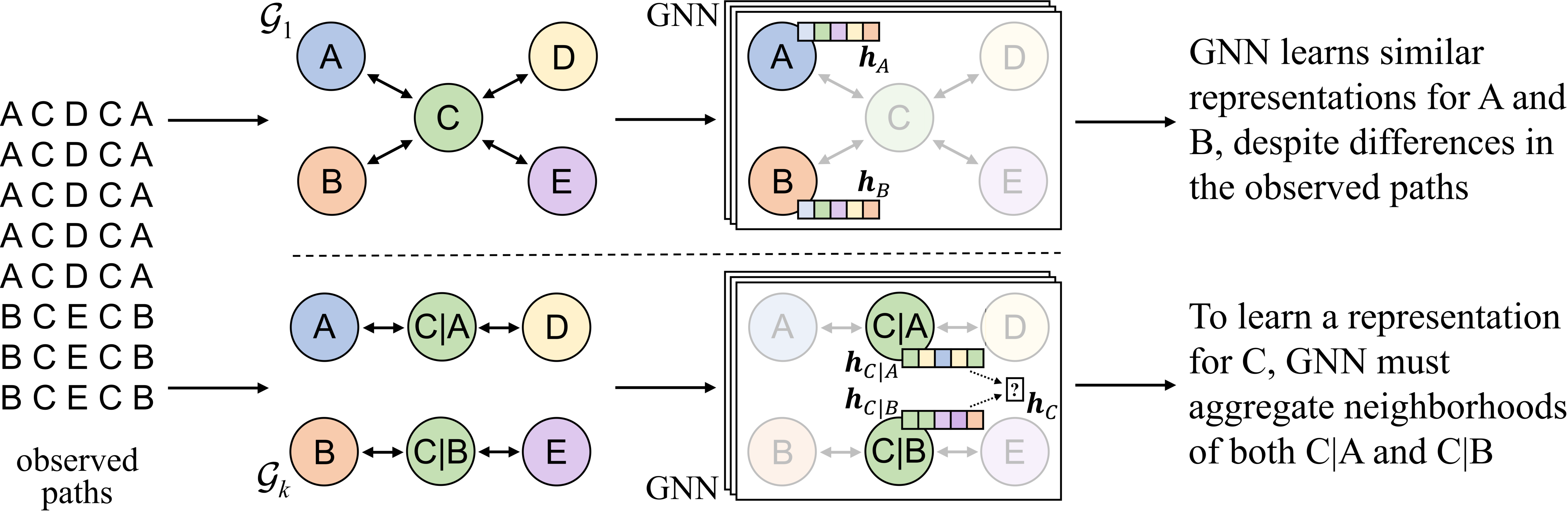}
    \caption{A toy example of challenges faced by GNNs in modeling systems with higher-order dependencies. A FON ($\myg_1$) underfits the higher-order dependencies in the observed paths. Consequently, a GNN will learn similar representations for A and B, since they share the same 2-hop neighborhood in $\myg_1$. A HON ($\myg_k$, with $k=2$ in this example) uses conditional nodes to encode higher-order dependencies. For example, node C$|$A represents the observed dependency that C only interacts with D when it also interacts with A (note that in real-world systems, $\myg_k$ rarely breaks the graph into multiple components). However, computing a representation for C then requires a GNN to aggregate multiple local neighborhoods. Colors depict node features.}
    \label{fig:toy}
\end{figure}

As Figure \ref{fig:toy} illustrates, the tendency of FONs to underfit has consequences for GNNs, which typically compute representations by recursively pooling features from each node's neighbors. In order to maximize GNN performance, we must ensure that local neighborhoods capture the true distribution of interactions in the system. To enable GNNs to utilize the additional information encoded in HONs, we propose a novel \textbf{Deep Graph Ensemble} (\prop), which uses independent GNNs to exploit variance in higher-order node neighborhoods and learn effective representations in graphs with higher-order dependencies. The \textbf{key contributions} of our work include:
\begin{enumerate}[noitemsep,topsep=0pt]
    \item We analyze the data-level challenges that fundamentally limit the ability of existing GNNs to learn effective models of systems with higher-order dependencies.
    \item We introduce the notion of neighborhood subspaces by showing that neighborhoods in a HON are analogous to feature subspaces of first-order neighborhoods. Borrowing from ensemble methods, we then propose \prop to exploit the variance in these subspaces.
    \item We experimentally evaluate \prop against eight state-of-the-art baselines on six real-world data sets with known higher-order dependencies, and show that, even with similar parameter budgets, \prop consistently outperforms baselines on semisupervised (node classification) and supervised (link prediction) tasks.\footnote{Code and 3 data sets are available at \url{https://github.com/sjkrieg/dge}.}
    \item We demonstrate that \prop's ability to train accurate and diverse classifiers is central to strong performance, and show that ensembling multiple GNNs with separate parameters is a consistent way to maximize the trade-off between accuracy and diversity.
\end{enumerate}


\section{Background and preliminaries} \label{sec:background}

\subsection{Higher-order networks} 
Let $\mathcal{S} = \{S_1, S_2, ..., S_n\}$ be a set of \textbf{observed paths} (e.g., flight itineraries, disease trajectories, or user clickstreams), where each $S_i = \langle s_1, s_2, ..., s_m \rangle $ is a sequence of \textbf{entities} (e.g., airports, diagnosis codes, or web pages). Let $\mathcal{A} = \bigcup \mathcal{S}$ denote the set of entities across all sequences. By using a graph to summarize $\mathcal{S}$, we can model the global function of each entity in the system and solve a number of useful learning problems. For example, we can predict disease function via node classification or interactions between airports using link prediction.\footnote{This is distinct from sequence models like transformers, which typically predict an entity's local function within a single sequence.} However, there is a large space of possible graphs that can represent $\mathcal{S}$. We consider two: a FON, and the HON introduced by \citet{xu2016representing}. In a FON $\myg_1 = (\myv_1, \mye_1)$, the node set $\myv_1 = \mathcal{A}$ (or, more generally, the mapping $f: \myv_1 \xrightarrow[ ]{}\mathcal{A}$ is bijective), and the edge set $\mye_1$ is the set of node pairs $(u, v) \in \myv_1 \times \myv_1 $ that are adjacent elements in at least one $S_i$. 

In a HON $\myg_k = (\myv_k, \mye_k)$ with order $k > 1$, each node is a sequence of entities $u' = \langle a'_1, ..., a'_{m-1}, a'_m \rangle$, where each $a'_i \in \myv_1$ and $m \leq k$. We define $a'_m$ as the \textbf{base node}, and in practice use the notation $u' = a'_m|a'_1, ..., a'_{m-1}$ to emphasize that each $u' \in \myv_k$ represents a base node whose current state is conditioned on a set of predecessors. Each node can have a different number of predecessors, and a conditional node with $m > 1$ is only created if the conditional distribution of paths it encodes sufficiently reduces the entropy of the graph \citep{saebi2020efficient}. This means that $\myv_k \subseteq \mathcal{A}^k$; or, more generally, the mapping $f: \myv_k \xrightarrow[ ]{}\mathcal{A}^k$ is injective but not necessarily bijective. We define $\Omega_u^k = \{u' \in \myv_k, a'_m = u\}$ as the \textbf{higher-order family} of $u$ (including $u$ itself), and call each $u' \in \Omega_u^k$ a \textbf{relative} of $u$ (in Figure 1, for example, C$|$A and C$|$B are the relatives of C). Like $\mye_1$, the edge set $\mye_k$ is the set of node pairs $(u, v) \in \myv_k \times \myv_k $ that are adjacent in at least one $S_i$. In both HONs and FONs, edges are directed such that $(u, v) \neq (v, u)$ and weighted via $w_k: \mye_k \xrightarrow[]{} \mathbb{R}_{\geq 0}$, where 0 indicates a missing edge. By creating conditional nodes, a HON can express higher-order interactions while remaining a graph, since each edge is still a 2-tuple. For example, consider that passengers who fly from Atlanta to Chicago are much more likely to fly back to Atlanta than to New York, and vice versa. A HON can encode this dependency by creating the conditional nodes ``Chicago$|$Atlanta'' and ``Chicago$|$New York'', which changes the topology and flow of information within local neighborhoods \citep{rosvall2014memory}. Choosing which conditional nodes and edges to create is a non-trivial problem; readers interested in more details can refer to \citet{krieg2020growhon}, who proposed the procedure that we used in this study.


\paragraph{Related graph-based models} The term ``higher-order'' is also used in the literature to refer to the analysis of polyadic structures within graphs \citep{benson2016higher}, as well GNNs that are able to distinguish these structures \citep{morris2019weisfeiler,li2020distance,schnake2021higher}. These studies rely on the same assumptions as other GNNs, i.e., that a graph is a sufficient approximation of the neighborhood structure. Despite similarities in terminology, HONs are primarily concerned with the question of initial representation (i.e., how should the graph be constructed?) rather than downstream analysis of an existing graph, and thus address a fundamentally different---though complementary---problem \citep{lambiotte2019networks}. HONs are also distinct from other formalisms like hypergraphs and simplicial complexes in that they encode conditional distributions that govern higher-order paths (via conditional nodes and directed, weighted edges), and therefore represent different kinds of systems \citep{battiston2020networks,porter2020nonlinearity,torres2021and,battiston2021physics}. Some very recent works have shown that aggregators based on paths (via random walks) can improve the expressiveness of GNNs \citep{eliasof2022pathgcn,jin2022raw}, but these still rely on the graph structure to guide path sampling. This further motivates the use of representations like HONs, which are designed to consistently and accurately encode higher-order paths for downstream learning tasks \citep{rosvall2014memory,xu2016representing,saebi2020efficient,krieg2020growhon}. To our knowledge, only one study has previously used GNNs with HONs \citep{jin2022graph}; however, as we discuss in Section \ref{sec:methods}, its proposed method has critical shortcomings.

\subsection{Graph neural networks}
A generic GNN computes a hidden vector representation for a node $u$ at timestamp $t$ according to:
\begin{equation} \label{eq:hu}
    \mathbf{h}_u^{(t)} = \textsc{\footnotesize{COMBINE}} \Big( \mathbf{h}_u^{(t-1)}, \textsc{\footnotesize{AGGREGATE}} \big(\{\mathbf{h}_v^{(t-1)}, v \in \mathcal{N}(u) \} \big) \Big),
\end{equation}
where $\mathcal{N}(u)$ is the neighborhood of $u$ in a graph $\myg$. What typically distinguishes GNNs is how they define $\textsc{\footnotesize{COMBINE}}$ and  $\textsc{\footnotesize{AGGREGATE}}$, and how they represent $\mathcal{N}(u)$ \citep{xu2018how,zhou2020graph,wu2020comprehensive}. By recursively pooling features via Eq. \ref{eq:hu}, GNNs implicitly construct higher-order neighborhoods across transitive paths. This allows nonadjacent nodes to share information if they are close in the graph \citep{li2018deeper,chen2020simple}; however, as we will demonstrate, when higher-order dependencies are present, assuming transitivity leads GNNs that are trained on FONs to generalize poorly.

In this work, we do not reformulate Eq. \ref{eq:hu} and are agnostic toward its particular implementation. We instead abstract GNN as a function that takes a single node $u$ as input and returns either the final hidden representation $\mathbf{h}^{(t)}_u$, or, in a supervised setting, a vector of predicted label probabilities $\hat{\mathbf{y}}_u$:
\begin{equation} \label{eq:gnn}
\gnn(u) = \mathbf{h}_u^{(t)} \quad \mathrm{or} \quad \gnn(u) = \hat{\mathbf{y}}_u.
\end{equation}
We always assume that $\myg$ contains initial node features $\{\mathbf{x}_u, \forall u \in \myv\}$ and that $\gnn(\cdot)$ is parameterized by weights $\theta$, but for simplicity we omit them from our notation.

\paragraph{GNN ensembles}
\citet{ho1998random} and \citet{breiman2001random} showed that training an ensemble of shallow learners on random subspaces could exploit variance in the feature space and improve performance, and \citet{dietterich2000ensemble} demonstrated that these ensembles are most potent when the predictions of their base classifiers are accurate and diverse. Deep ensembles have typically been used with random weight initializations to improve uncertainty estimation and robustness \citep{lakshminarayanan2017simple,fort2019deep,wasay2020more}, which has benefited GNNs via mechanisms like multi-head attention \citep{velivckovic2018graph,brody2021attentive,hou2021graph}, but very few works have directly explored ensembles of GNNs. Some recent exceptions have suggested that ensembling subgraphs could benefit GNNs \citep{zeng2021decoupling,tang2021graph,lin2022robust}.
\section{Deep graph ensembles for higher-order networks} \label{sec:methods}

\begin{figure}[t]
    \centering
    \includegraphics[width=\textwidth]{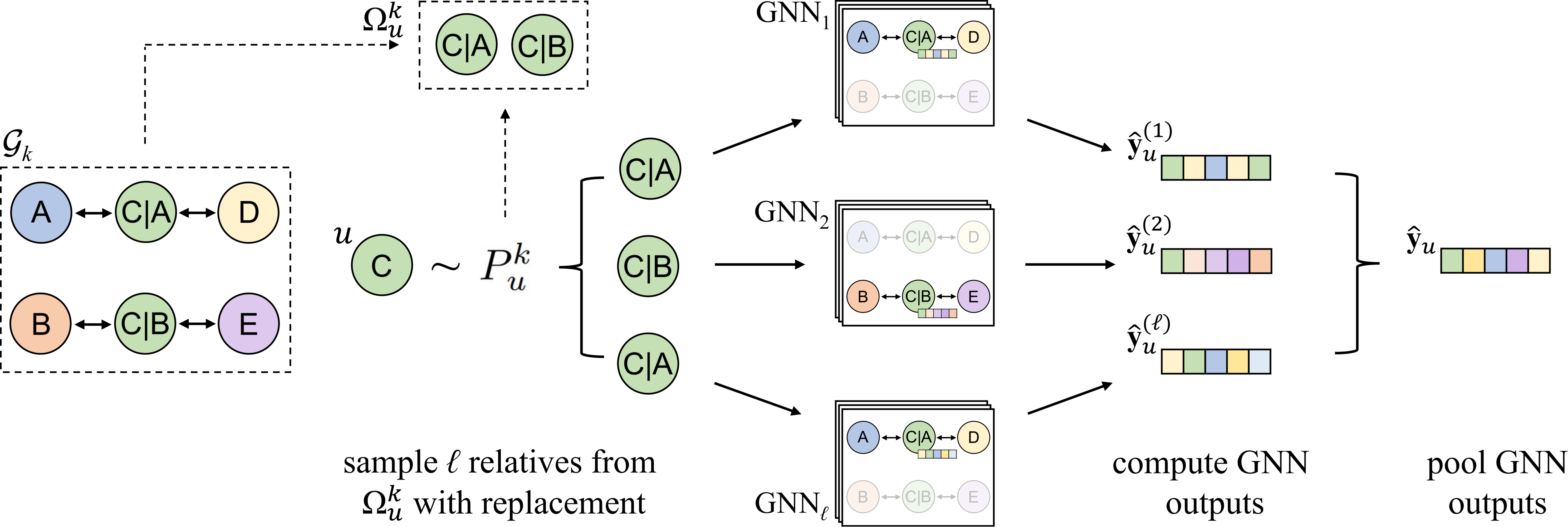}
    \caption{Overview of \prop. Given a HON $\myg_k$, \prop computes outputs for each base node $u$ by a) resampling relatives from $u$'s higher-order family $\Omega_u^k$ via a sampling distribution $P_u^k$, b) computing outputs for each sampled relative using independent GNN modules, and c) pooling the outputs.}
    \label{fig:overview}
\end{figure}

\subsection{Why ensembles?} \label{sec:whyens}
There are a number of design challenges (\textbf{CHs}) we must address in order to realize the joint potential of GNNs and HONs. In a HON, entities are represented by a non-fixed number of conditional nodes (\textbf{CH1}); for example, in Figure 1, C is represented by two nodes but A, B, D, and E are each only represented by one node. Conditional nodes typically have different neighborhoods (\textbf{CH2}); for example, in Figure 1, C$|$A and C$|$B have different neighbors. Further, they may vary in importance (i.e., degree) in the graph (\textbf{CH3}) \citep{xu2016representing,saebi2020efficient}. To address these challenges, one intuitive idea is to reformulate Eq. \ref{eq:gnn} so that it computes a representation for $u$ by sampling neighbors from any of $u$'s relatives. However, this fails to address CH2 because a GNN would aggregate the samples without considering differences between relatives. Another idea is to compute representations for each relative separately, then pool them via a permutation-invariant function like an elementwise \textsc{\footnotesize{MEAN}}, as proposed for HO-GNN by \citet{jin2022graph}. But this does not address CH3, since all relatives would contribute equally to the final representation. Moreover, if we assume that all relatives in a higher-order family share the same (or similar) features, this solution will overrepresent features associated with larger higher-order families. In order to propose a method that comprehensively addresses these challenges, we first consider the following relationship.

\newtheorem{theorem}{Theorem}
\begin{theorem} \label{th:subspace}
Let $\myg_1$ and $\myg_k$ be a FON and HON, respectively, both constructed from the same input $\mathcal{S}$. Let $\mathcal{N}_1(u)$ and $\mathcal{N}_k(u)$ denote the neighborhoods of any node $u$ in $\myg_1$ and $\myg_k$, respectively. Let $\textsc{\footnotesize{AGGREGATE}}(\cdot)$ represent any symmetric neighborhood aggregation function. If $u \in \myv_1$ and $u' \in \Omega_u^k$, then $\textsc{\footnotesize{AGGREGATE}}(\mathcal{N}_k(u'))$ is a biased estimator of $\textsc{\footnotesize{AGGREGATE}}(\mathcal{N}_1(u))$. 
\end{theorem}

We prove Theorem \ref{th:subspace} in Appendix \ref{sec:proof}. Intuitively, we observe that HONs are constructed such that $\mathcal{N}_k(u') \subseteq \mathcal{N}_1(u)$, and $u'$ only exists in $\myg_k$ if the expectation of a random walker differs substantially (measured via KL-divergence) from $u$ in $\myg_1$ \citep{saebi2020efficient,krieg2020growhon}. Consequently, these differences in neighborhood structure will shift the expectation of the features gathered by $\textsc{\footnotesize{AGGREGATE}}(\mathcal{N}_k(u'))$. A typical strategy for training a single GNN would involve attempting to eliminate this variance via some sampling method on the graph. We instead take an ensemble approach and propose to regularize the model via multiple GNNs that err in different ways. Toward this end, and inspired by feature subspace methods \citep{ho1998random,breiman2001random}, we call $\mathcal{N}_k(u')$ a \textbf{neighborhood subspace} of $\mathcal{N}_1(u)$.

\paragraph{Remark} Our result from Theorem \ref{th:subspace} also relates to the expressiveness of an aggregator over neighborhoods in $\myg_k$. Consider two nodes $u, v\in \myv_1$ whose rooted subgraphs (i.e., neighborhoods) are non-isomorphic but are not distinguishable by the aggregator in $\myg_1$. As long as there exists some $u' \in \Omega_u^k$ (excluding $u$ itself), then we know that, since it is a biased estimator, the aggregator can distinguish $u'$ from $u$ in $\myg_k$. It follows transitively that the aggregator can also distinguish $u'$ from $v$. It is possible that there also exists some $v' \in \Omega_v^k$ that cannot be distinguished from $u'$, but this would be extremely unlikely to occur over all pairs of relatives in $\Omega_u^k$ and $\Omega_v^k$. We can thus improve the expressiveness of the model by allowing the information from nodes in $\Omega_u^k$ to contribute to the final representation for each $u$.

Guided by these observations, we propose to address the CHs outlined above by training an ensemble of GNNs $\{\gnn_1, \gnn_2, ..., \gnn_{\ell} \}$. Given a set of training nodes $D \subseteq \myv_1$, we generate bootstraps $\{D^{(1)}, D^{(2)}, ..., D^{(\ell)} \}$ subject to the constraint that $\vert D^{(i)} \bigcap \Omega_u^k \vert = 1$ for all $u \in D$ and $i \leq \ell $ (i.e., each bootstrap contains exactly one relative of each training node). This constraint allows us to avoid the feature overrepresentation problem, address CH1 by sampling with replacement, and address CH2 by training each $\gnn_i$ on different neighborhood subspaces as represented in $D^{(i)}$. To solve CH3, we weight the sampling probability for each relative $u'$ according to the normalized out-degree of its higher-order family:
\begin{equation} \label{eq:samp}
P_u^k (u') = \frac{\textsc{\footnotesize{outdeg}}_k(u')}{\sum_{v' \in \Omega_u^k} \textsc{\footnotesize{outdeg}}_k(v')},
\end{equation}
where $u \in D$ and $\textsc{\footnotesize{outdeg}}_k(u')$ is the weighted out-degree of $u'$ in $\myg_k$. Because weighted out-degree in $\myg_k$ is the frequency with which $u'$ appears in $\mathcal{S}$ \citep{krieg2020growhon}, it is a natural measure of the importance of $u'$ with respect to the rest of the higher-order family $\Omega_u^k$. If $D \subseteq \mye_1$ consists of node pairs for an edge task, we modify Eq. \ref{eq:samp} slightly. For each edge $(u,v) \in D$, we resample a single pair of relatives $(u', v')$ with probability according to the normalized weights of all edges between relatives of $u$ and $v$:
\begin{equation} \label{eq:sampedges}
P_{u,v}^k (u', v') = \frac{w_k(u', v')}{\sum_{(u'', v'') \in \Omega_u^k \times \Omega_v^k} \big(w_k(u'', v'')\big)}.
\end{equation}


\subsection{Training and inference} \label{sec:ensimp}
Let $D^{(i)}_u$ denote the relative of $u$ that was sampled for the $i^{th}$ bootstrap. We consider a supervised or semisupervised setting, in which our goal is to predict class probabilities $\mathbf{\hat{y}}_u$ for each $u \in D$ such that some loss is minimized w.r.t. ground truth $\mathbf{y}_u$. We propose three methods for computing $\mathbf{\hat{y}}_u$:
\begin{subequations}
\begin{equation} \label{eq:propa}
\hat{\mathbf{y}}_u = \sigma \Big( \textsc{\footnotesize{CONCAT}} \big( \big\{ \gnn_i\
(D^{(i)}_u), \forall i \leq \ell\big\} \big)^{\top} \cdot \mathbf{W} \Big) \quad \textrm{and} \quad \gnn_i(u') = \mathbf{h}_{u'}^{(i,t)},
\end{equation} 
\begin{equation} \label{eq:propb}
\hat{\mathbf{y}}_u = \sigma \Big( \textsc{\footnotesize{MEAN}} \big( \big\{\gnn_i(D^{(i)}_u), \forall i \leq \ell\big\} \big)^{\top} \cdot \mathbf{W} \Big) \quad \textrm{and} \quad \gnn_i(u') = \mathbf{h}_{u'}^{(i,t)},
\end{equation} 
\begin{equation} \label{eq:propc}
\hat{\mathbf{y}}_u = \textsc{\footnotesize MEAN} \Big( \big\{ \gnn_i(D^{(i)}_u), \forall i \leq \ell\big\} \Big) \quad \textrm{and} \quad \gnn_i(u') = \hat{\mathbf{y}}_{u'}^{(i)},
\end{equation}
\end{subequations}

where $\sigma$ is a non-linear activation, $\textsc{\footnotesize{CONCAT}}$ is vector concatenation, $\textsc{\footnotesize{MEAN}}$ is elementwise mean, $\mathbf{x}^{\top}$ is the transpose of $\mathbf{x}$, $\mathbf{h}_{u'}^{(i,t)} \in \mathbb{R}^d$ represents the hidden state of node $u'$ in the $t^{th}$ (final) layer of $\gnn_i$, and $d$ is the number of hidden units (we assume $d$ is fixed for all $\gnn_i$). Using $c$ to denote the number of classes, for Eq. \ref{eq:propa} we have $\mathbf{W} \in \mathbb{R}^{d\ell \times c}$, and for Eq. \ref{eq:propb} we have $\mathbf{W} \in \mathbb{R}^{d \times c}$. We refer to Eq. \ref{eq:propa} as \textbf{\propa}, Eq. \ref{eq:propb} as \textbf{\propb}, and Eq. \ref{eq:propc} as \textbf{\propc}.

In \propa~and \propb~each GNN outputs hidden representations, which are concatenated or pooled, respectively, before computing the logits. This means that they can be trained in end-to-end fashion as a single neural network with parallel but independent GNN modules. During a forward pass, each $\gnn_i$ only computes representations for the nodes in $D^{(i)}$. In \propc, on the other hand, each GNN outputs class probabilities, which means that each GNN is trained independently and their predictions are simply averaged to compute the final probabilities. We also designed an attention-based pooling method, but found that it did not generalize well (Appendix \ref{sec:additionalpooling}).

One important question is whether all $\gnn_i$ should share parameters. In other words, is an ensemble necessary, or is it sufficient to use single, more complex model \citep{abe2022deep}? We evaluated this question experimentally, and use \textbf{\propa}*, \textbf{\propb}*, and \textbf{\propc}* to denote shared-parameter variants of Eqs. \ref{eq:propa}, \ref{eq:propb}, and \ref{eq:propc}, respectively. For \propa* and \propb*, we adjusted our training procedure so that, during backpropagation, each parameter was updated once according to its contribution to the summed loss across all $D^{(i)}$. For \propc*, this meant that each $\gnn_i$ was essentially pretrained on $D^{(i-1)}$. Since each $\gnn_i$ shares parameters, none of these variants are true ensembles. Instead, they are single models that synthesize representations for conditional nodes via a $\textsc{\footnotesize{READOUT}}$ function \citep{wu2020comprehensive} on each higher-order family. \propb* is similar to HO-GNN \citep{jin2022graph}, which uses a single GNN and computes a representation for each $u$ via the mean of all its relatives (rather than a weighted sample) in $\Omega_u^k$. We also considered one final variant, \textbf{\propd}*, which does not use a fixed set of bootstraps for training. Instead, it uses Eq. \ref{eq:samp} to sample a new set of relatives for each batch. Then, during inference, we use the same procedure as \propc: resample $\ell$ relatives for each node and compute their outputs via Eq. \ref{eq:propc}. \propd* thus drops any resemblance to an ensemble, instead addressing CH1, CH2, and CH3 entirely via batch sampling.

Figure \ref{fig:overview} summarizes the components of \prop: resampling $\ell$ relatives for each entity via Eq. \ref{eq:samp}, computing a node representation for each sampled relative via Eq. \ref{eq:gnn}, and pooling the computed representations via Eq. \ref{eq:propa}, \ref{eq:propb}, or \ref{eq:propc}. In general, \prop's computational cost is linear with the cost of the base GNN and the ensemble size $\ell$, since we are essentially constructing $\ell$ copies of that GNN (or, in the case of shared parameters, repeating $\ell$ forward passes per example). For a given node $u$, the additional cost of sampling and pooling $\ell$ relatives is $O(\ell \ |\Omega_u^k|$) and $O(\ell)$, respectively, which are trivial compared to the costs of Eq. \ref{eq:gnn} \citep{wu2020comprehensive}. Additionally, the one-time cost of constructing $\myg_k$ increases linearly with $k$ \citep{krieg2020growhon}. Since we used $\myg_2$ in our experiments, there was marginal overhead for graph construction as compared to $\myg_1$. 
\section{Experimental results and discussion} \label{sec:results}

\subsection{Experimental setup} \label{sec:setup}

\begin{table}[t]
\caption{Summary of graphs used in node classification and link prediction experiments.}
\label{tab:data}
\centering
\begin{tabular}{lcccccccc}
\toprule
     Name & $|S|$ & $|\myv_1|$ & $|\mye_1|$ & $|\myv_2|$ & $|\mye_2|$ & \# classes & $\mathcal{H}(\myg_1)$ & $\mathcal{H}(\myg_2 )$\\
     \cmidrule(lr){1-9}
Air  & 17.1m & 416 & 13,735 & 7,461 & 236,806 & 10 & 0.351 & 0.288 \\
T2D  & 0.8m & 908 & 314,352 & 5,462 & 367,530 & 16 & 0.099 & 0.099 \\
Wiki & 76.2k & 4,179 & 70,662 & 5,584 & 84,806 & 10 & 0.366 & 0.375 \\
Mag   & 3.8m & 4,079 & 1,873,279 & 9,568 & 1,957,602 & 4 & 0.323 & 0.336 \\
Mag+  & 8.1m & 17,428 & 5,098,787 & 192,204 & 7,980,026 & 6 & 0.292 & 0.310 \\
Ship  & 54.9k & 5,586 & 369,952 & 8,586 & 422,426 & 38 & 0.477 & 0.489 \\
\bottomrule
\end{tabular}
\end{table}

We used $\textsc{\footnotesize{GrowHON}}$ \citep{krieg2020growhon} to construct FONs ($\myg_1$) and HONs with $k=2$ ($\myg_2$) for six real-world data sets with known higher-order dependencies: flight itineraries for airline passengers in the United States (\textbf{Air}) \citep{rosvall2014memory}, disease trajectories for type 2 diabetes patients in Indiana (\textbf{T2D}) \citep{krieg2020higher}, clickstreams of users playing the Wikispeedia game (\textbf{Wiki}) \citep{west2009wikispeedia}, readership trajectories for a large online magazine (\textbf{Mag} and a larger version, \textbf{Mag+}, for node classification only) \citep{wang2020calendar}, and global shipping routes (\textbf{Ship}) \citep{saebi2020higher}. Table \ref{tab:data} summarizes their key characteristics, including average homophily $\mathcal{H}$ (see Appendix \ref{sec:nodechar} for details). We discuss additional details and preprocessing steps in Appendix \ref{sec:datasum}. 

We evaluated several baselines, including GCN \citep{kipf2017semi}, GAT \citep{velivckovic2018graph}, GraphSAGE \citep{hamilton2017inductive}, GIN \citep{xu2018how} and GATv2 \citep{brody2021attentive}, and SEAL \citep{zhang2018link} (link prediction only). Other noteworthy baselines are GraphSAINT \citep{zeng2019graphsaint} which samples a different subgraph for each training iteration \citep{hu2020open}; GCNII \citep{chen2020simple}, which uses residual connections to address over-smoothing; and PathGCN \citep{eliasof2022pathgcn}, which learns spatial operators on paths sampled via random walks. We also evaluated two baselines designed specifically for HONs: HONEM, a matrix factorization method \citep{saebi2020honem}, and HO-GNN \citep{jin2022graph}. All baselines used $\myg_1$ as input (we also evaluated each baseline using $\myg_2$ as input, details are in Appendix \ref{sec:supresults}). We manually tuned each model (details in Appendix \ref{sec:hyperparams}). For \prop, unless noted otherwise, we fixed $\ell = 16$ and used the mean-pooling variant of GraphSAGE as the base GNN, since a) it performed reasonably well as a baseline on all six data sets, and b) its sample-and-aggregate procedure intuitively complements the relative sampling and pooling used by \prop. We used Python 3.7.3 and Tensorfow 2.4.1 for all experiments, and utilized Stellargraph 1.2.1 \citep{StellarGraph} for the implementation of \prop.

\subsection{Experimental results}

\paragraph{Node classification and link prediction}
\begin{table}[t]
\caption{Node classification results (micro F1). Bold font indicates the best result for each data set.}
\label{tab:nc}
\centering
\begin{tabular}{lcccccc}
\toprule
Model & Air & T2D & Wiki & Mag & Mag+ & Ship \\ 
\cmidrule(lr){1-7}
GCN             & 0.818 \scriptsize $\pm$ 0.03 & 0.480 \scriptsize $\pm$ 0.02 & 0.643 \scriptsize $\pm$ 0.01 & 0.796 \scriptsize $\pm$ 0.01 & 0.710 \scriptsize $\pm$ 0.01 & 0.746  \scriptsize $\pm$ 0.01 \\
GCNII           & 0.839  \scriptsize $\pm$ 0.05 & 0.511 \scriptsize $\pm$ 0.02 & 0.654 \scriptsize $\pm$ 0.02 & 0.801 \scriptsize $\pm$ 0.01 & 0.712 \scriptsize $\pm$ 0.01 & 0.770 \scriptsize $\pm$ 0.01 \\
GAT             & 0.804 \scriptsize $\pm$ 0.03 & 0.282 \scriptsize $\pm$ 0.10 & 0.639 \scriptsize $\pm$ 0.02 & 0.487 \scriptsize $\pm$ 0.06 & 0.511 \scriptsize $\pm$ 0.02 & 0.745 \scriptsize $\pm$ 0.02 \\
GATv2           & 0.838 \scriptsize $\pm$ 0.03 & 0.292 \scriptsize $\pm$ 0.07 & 0.643 \scriptsize $\pm$ 0.03 & 0.495 \scriptsize $\pm$ 0.05 & 0.527 \scriptsize $\pm$ 0.02 & 0.747  \scriptsize $\pm$ 0.01 \\
GraphSAGE       & 0.781 \scriptsize $\pm$ 0.04 & 0.654 \scriptsize $\pm$ 0.04 & 0.625 \scriptsize $\pm$ 0.02 & 0.808 \scriptsize $\pm$ 0.02 & 0.689 \scriptsize $\pm$ 0.01 & 0.796  \scriptsize $\pm$ 0.01 \\
GIN             & 0.745 \scriptsize $\pm$ 0.02 & 0.673 \scriptsize $\pm$ 0.04 & 0.636 \scriptsize $\pm$ 0.02 & 0.826 \scriptsize $\pm$ 0.02 & 0.722 \scriptsize $\pm$ 0.00 & 0.801  \scriptsize $\pm$ 0.01 \\
GraphSAINT      & 0.802 \scriptsize $\pm$ 0.02 & 0.600 \scriptsize $\pm$ 0.07 & 0.664 \scriptsize $\pm$ 0.01 & 0.821 \scriptsize $\pm$ 0.02 & 0.719 \scriptsize $\pm$ 0.01 & 0.803 \scriptsize $\pm$ 0.01 \\
PathGCN & 0.846 \scriptsize $\pm$ 0.02 & 0.523 \scriptsize $\pm$ 0.03 & 0.616 \scriptsize $\pm$ 0.03 & 0.735 \scriptsize $\pm$ 0.02 & 0.622 \scriptsize $\pm$ 0.01 & 0.781 \scriptsize $\pm$ 0.01 \\
HONEM           & 0.805 \scriptsize $\pm$ 0.04 & 0.566 \scriptsize $\pm$ 0.02 & 0.588 \scriptsize $\pm$ 0.01 & 0.728 \scriptsize $\pm$ 0.02 & 0.620 \scriptsize $\pm$ 0.01 & 0.750 \scriptsize $\pm$ 0.01 \\
HO-GNN & 0.822 \scriptsize $\pm$ 0.02 & 0.436 \scriptsize $\pm$ 0.03 & 0.581 \scriptsize $\pm$ 0.01 & 0.785 \scriptsize $\pm$ 0.01 & 0.702 \scriptsize $\pm$ 0.01 & 0.787 \scriptsize $\pm$ 0.01 \\
\cmidrule(lr){1-7}
\propa          & 0.825 \scriptsize $\pm$ 0.04 & 0.501 \scriptsize $\pm$ 0.06 & 0.615 \scriptsize $\pm$ 0.02 & 0.790 \scriptsize $\pm$ 0.02 & 0.681 \scriptsize $\pm$ 0.04 & 0.828 \scriptsize $\pm$ 0.01 \\
\propa*         & 0.810 \scriptsize $\pm$ 0.04 & 0.439 \scriptsize $\pm$ 0.03 & 0.577 \scriptsize $\pm$ 0.02 & 0.761 \scriptsize $\pm$ 0.02 & 0.642 \scriptsize $\pm$ 0.01 & 0.809 \scriptsize $\pm$ 0.01 \\
\propb          & 0.839 \scriptsize $\pm$ 0.03 & 0.735 \scriptsize $\pm$ 0.03 & 0.671 \scriptsize $\pm$ 0.01 & 0.860 \scriptsize $\pm$ 0.01 & 0.722 \scriptsize $\pm$ 0.01 & 0.808 \scriptsize $\pm$ 0.01 \\
\propb*         & \textbf{0.865} \scriptsize $\pm$ 0.02 & 0.555 \scriptsize $\pm$ 0.07 & 0.599 \scriptsize $\pm$ 0.04 & 0.775 \scriptsize $\pm$ 0.01 & 0.671 \scriptsize $\pm$ 0.01 & 0.767 \scriptsize $\pm$ 0.02 \\
\propc          & 0.856 \scriptsize $\pm$ 0.02 & \textbf{0.770} \scriptsize $\pm$ 0.04 & \textbf{0.681} \scriptsize $\pm$ 0.00 & \textbf{0.871} \scriptsize $\pm$ 0.01 & \textbf{0.769 \scriptsize $\pm$ 0.01} & \textbf{0.840 \scriptsize $\pm$ 0.01} \\
\propc*         & 0.766 \scriptsize $\pm$ 0.04 & 0.719 \scriptsize $\pm$ 0.04 & 0.644 \scriptsize $\pm$ 0.02 & 0.841 \scriptsize $\pm$ 0.02 & 0.739 \scriptsize $\pm$ 0.01 & 0.825 \scriptsize $\pm$ 0.01 \\
\propd*         & 0.764 \scriptsize $\pm$ 0.03 & 0.646 \scriptsize $\pm$ 0.01 & 0.623 \scriptsize $\pm$ 0.01 & 0.818 \scriptsize $\pm$ 0.01 & 0.742 \scriptsize $\pm$ 0.00 & 0.812 \scriptsize $\pm$ 0.01 \\
\bottomrule
\scriptsize{* shared parameters}
\end{tabular}
\end{table}

\begin{table}[t]
\caption{Link prediction results (AUPRC). Bold font indicates the best result for each data set.}
\label{tab:lp}
\centering
\begin{tabular}{lccccc}
\toprule
Model & Air & T2D & Wiki & Mag & Ship \\
\cmidrule(lr){1-6}
\multirow{2}{*}{Best baseline} & 0.818 \scriptsize $\pm$ 0.02 & 0.818 \scriptsize $\pm$ 0.00 & 0.834 \scriptsize $\pm$ 0.01 & 0.879 \scriptsize $\pm$ 0.00 & 0.887 \scriptsize $\pm$ 0.01 \\
                                & \footnotesize (SEAL) & \footnotesize  (HONEM) & \footnotesize (SEAL) & \footnotesize  (HO-GNN) & \footnotesize (SEAL) \\
\cmidrule(lr){1-6}
\propa          & \textbf{0.886} \scriptsize $\pm$ 0.01 & 0.815 \scriptsize $\pm$ 0.01 & 0.774 \scriptsize $\pm$ 0.01 & 0.802 \scriptsize $\pm$ 0.00 & \textbf{0.910} \scriptsize $\pm$ 0.00 \\
\propa* & 0.856 \scriptsize $\pm$ 0.01 & 0.779 \scriptsize $\pm$ 0.01 & 0.712 \scriptsize $\pm$ 0.02  & 0.769 \scriptsize $\pm$ 0.01 & 0.904 \scriptsize $\pm$ 0.00 \\
\propb          & 0.851 \scriptsize $\pm$ 0.02 & \textbf{0.920} \scriptsize $\pm$ 0.00 & 0.838 \scriptsize $\pm$ 0.03 & 0.913 \scriptsize $\pm$ 0.00 & 0.898 \scriptsize $\pm$ 0.00 \\
\propb* & 0.845 \scriptsize $\pm$ 0.01 & 0.901 \scriptsize $\pm$ 0.00 & 0.802 \scriptsize $\pm$ 0.06 & 0.898 \scriptsize $\pm$ 0.00 & 0.815 \scriptsize $\pm$ 0.00 \\
\propc          & \textbf{0.887} \scriptsize $\pm$ 0.00 & 0.907 \scriptsize $\pm$ 0.00 & \textbf{0.876} \scriptsize $\pm$ 0.01 & \textbf{0.921} \scriptsize $\pm$ 0.00 & 0.895 \scriptsize $\pm$ 0.00 \\
\propc* & 0.862 \scriptsize $\pm$ 0.02 & 0.894 \scriptsize $\pm$ 0.00 & 0.856 \scriptsize $\pm$ 0.00 & 0.916 \scriptsize $\pm$ 0.00 & 0.891 \scriptsize $\pm$ 0.00 \\
\propd* & 0.853 \scriptsize $\pm$ 0.03 & 0.871 \scriptsize $\pm$ 0.01 & 0.830 \scriptsize $\pm$ 0.01 & 0.887 \scriptsize $\pm$ 0.00 & 0.865 \scriptsize $\pm$ 0.00 \\
\bottomrule
\scriptsize{* shared parameters}
\end{tabular}
\end{table}

On the node classification task (Table \ref{tab:nc}), \propc~outperformed all other methods on five data sets  and was second to \propb* on Air. \propb~performed only slightly worse than \propc~in all cases, but \propa~did not generalize well and performed poorly in all cases except Ship. This suggests that our relative sampling procedure was effective in conjunction with ensembling (\propc), but not as effective at regularizing a single fully-connected network against the variance in $\myg_k$. Since concatenation depends heavily on the position of each relative in the vector representation, \propa~overfit on nodes that had many relatives. Sharing parameters typically reduced performance for each \prop variant, excepting \propb* on Air. We discuss this observation in Section \ref{sec:ensdiv}. The differences between HO-GNN and \prop emphasize the importance of accounting for both the variance in neighborhood subspaces in $\myg_k$ as well as the differences in importance between relatives. Some baselines were occasionally competitive: GCNII, GATv2, and PathGCN performed relatively well on Air, and GCNII and GraphSAINT performed relatively well on Wiki. However, on Mag and especially T2D, \propc~outperformed all baselines by a significant margin. PathGCN and the full-batch models struggled on T2D, likely due to its high density---meaning that there are many transitive and false positive paths in $\myg_1$. Low homophily may also have contributed to their poor performance \citep{zhu2020beyond} (see discussion in Appendix \ref{sec:nodechar}). GAT and GATv2 performed especially poorly on the dense graphs, likely because they compute attention weights (instead of using the given edge weights) and were thus more susceptible to overfitting. The link prediction results, summarized in Table \ref{tab:lp}, tell a similar story. Because of its strong performance, we focus the remainder of our analysis on \propc. For details on training time and convergence for all models, please see Appendix \ref{sec:traintime}.

\begin{figure}[t]
    \centering
    \includegraphics[width=\linewidth]{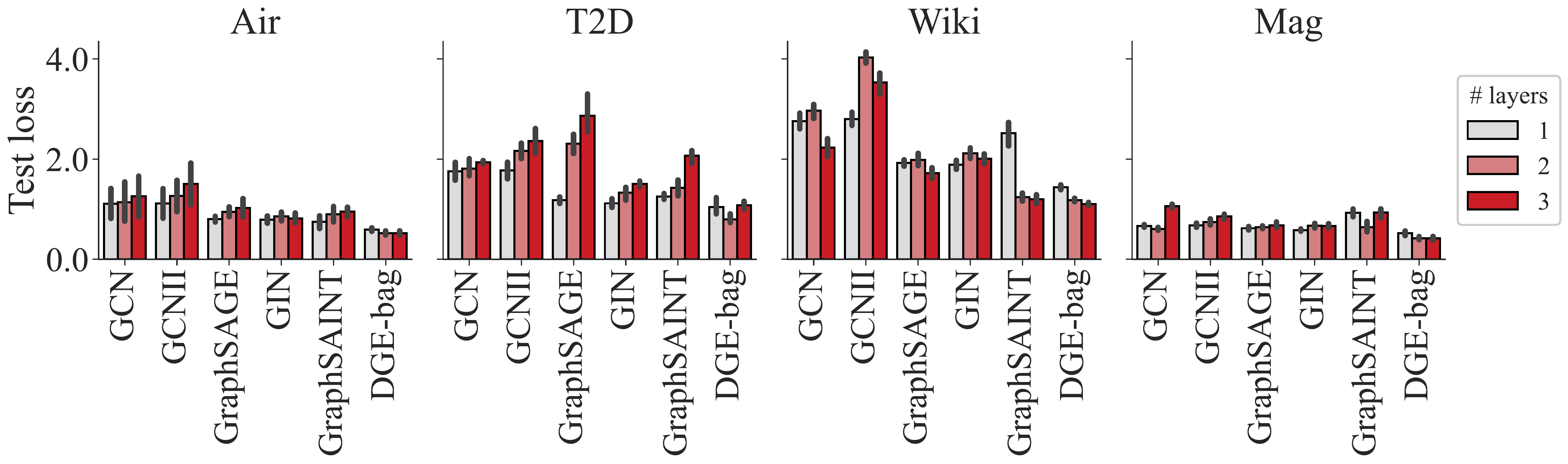}
    \caption{Node classification loss as a function of model depth. Error bars are standard deviation.}
    \label{fig:depthloss}
\end{figure}

\paragraph{Model size} \label{sec:modeldepth}
For the baselines, increasing the number of GNN layers typically increased generalization error (Figure \ref{fig:depthloss}). For \propc, the test loss decreased with increased model depth in all but one case (layer 3 on T2D). These results support the findings of \citet{lambiotte2019networks} that transitively inferring paths in a FON cannot account for higher-order dependencies, and our hypothesis that GNNs trained on $\myg_1$ will overfit on non-existent paths. The exception to this pattern was Wiki, on which many models performed best with 3 layers. This difference is perhaps because Wiki was the sparsest graph, meaning that the likelihood of sampling a false positive path is relatively low. However, \propc~still produced the lowest test error at all depths.

In order to ensure that \prop's performance was not simply due to using a more expensive model, we compared \propc~to the strongest baselines at several parameter budgets (Table \ref{tab:modelwidth}; results for Wiki and Mag are available in Appendix \ref{sec:supresults}). \propc~underperformed with the smallest budgets, since each base learner was too weak and underfit. We found that sharing parameters (\propc*) resolved this issue and produced strong results on T2D, Wiki, and Mag. However, with a moderate parameter budget, \propc~consistently outperformed other models. Again, for most baselines, increasing model depth simply caused caused them to overfit and decreased performance.

\begin{table}[t]
\caption{Node classification results (mean micro F1 for 5-fold cross validation) under various parameter budgets. Bold font indicates the best result for each budget and data set.}
\label{tab:modelwidth}
\centering
\begin{tabular}{lccccccccccc}
\toprule
& & \multicolumn{5}{c}{Total parameters (Air)} & \multicolumn{5}{c}{Total parameters (T2D)} \\
Model & Layers & 10k & 50k & 100k & 500k & 1m & 10k & 50k & 100k & 500k & 1m \\
\cmidrule(lr){1-2} \cmidrule(lr){3-7} \cmidrule(lr){8-12}
GCNII  & 2 & \textbf{0.78} & \textbf{0.79} & 0.80 & 0.80 & 0.80 & 0.43 & 0.47 & 0.47 & 0.49 & 0.50 \\
        & 4 & 0.77 & 0.78 & 0.78 & 0.79 & 0.79 & 0.43 & 0.47 & 0.50 & 0.51 & 0.51 \\
        & 8 & 0.74 & 0.76 & 0.76 & 0.76 & 0.76 & 0.45 & 0.49 & 0.49 & 0.51 & 0.51 \\
GATv2  & 2 & 0.66 & 0.73 & 0.80 & 0.82 & 0.83 & 0.11 & 0.14 & 0.17 & 0.18 & 0.18 \\
         & 4 & 0.49 & 0.58 & 0.62 & 0.71 & 0.75 & 0.10 & 0.12 & 0.14 & 0.14 & 0.15\\
GIN    & 2 & 0.74 & 0.74 & 0.75 & 0.75 & 0.75 & 0.54 & 0.64 & 0.67 & 0.66 & 0.67 \\
        & 4 & 0.63 & 0.72 & 0.73 & 0.73 & 0.73 & 0.36 & 0.38 & 0.38 & 0.44 & 0.46 \\
GraphSAINT & 2 & 0.71 & 0.75 & 0.75 & 0.76 & 0.77 & 0.45 & 0.56 & 0.59 & 0.60 & 0.61 \\
        & 4 & 0.60 & 0.64 & 0.68 & 0.74 & 0.74 & 0.25 & 0.27 & 0.28 & 0.28 & 0.28 \\
\cmidrule(lr){1-2} \cmidrule(lr){3-7} \cmidrule(lr){8-12}
DGE-bag                 & 2 & 0.68 & \textbf{0.79} & \textbf{0.81} & \textbf{0.83} & \textbf{0.86} & 0.47 & 0.66 & \textbf{0.73} & \textbf{0.75} & \textbf{0.77} \\
DGE-bag*                & 2 & 0.70 & 0.76 & 0.80 & 0.81 & 0.81 & \textbf{0.65} & \textbf{0.70} & 0.70 & 0.71 & 0.71 \\
\bottomrule
\end{tabular}
\end{table}

\subsection{Ensemble diversity and parameter sharing} \label{sec:ensdiv}

\begin{figure}[t]
    \centering
    \begin{subfigure}{\linewidth}
    \includegraphics[width=\linewidth]{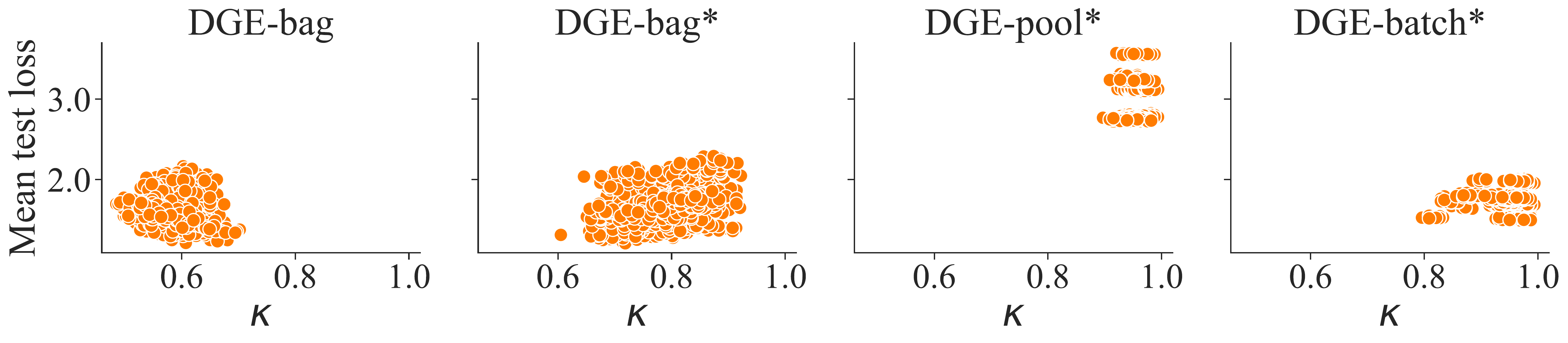}
    \end{subfigure}
    \caption{Mean node classification loss for all pairs of the $\ell=16$ base GNNs within each testing fold on T2D, plotted as a function of Cohen's kappa (lower values indicate lower agreement). Each point represents one pair of GNNs in the ensemble. All plots contain the same number of points.}
    \label{fig:kappas}
\end{figure}

The variants of \prop that used separate parameters generally performed best. To understand this observation, we draw from prior work on ensembles, which has established that ensembles are most potent when the individual classifiers have low error and high disagreement \citep{dietterich2000ensemble}. As Figure \ref{fig:kappas} shows, \propc~consistently produced classifiers that were both diverse and accurate. The shared-parameter variants often produced classifiers that were either diverse or accurate, but not both. We observed similar results for Air, Wiki, and Mag (Appendix \ref{sec:supresults}). These results reflect node classification performance (Table \ref{tab:nc}) and support several important conclusions. First, the high disagreement in \propc~demonstrates that treating higher-order relatives as neighborhood subspaces successfully introduced variance into the model. Second, the low mean error for \propc~demonstrates that ensembling GNNs with separate parameters effectively captured and exploited this variance. Third, sharing parameters (i.e., using a single, more complex model) did not achieve the same effect. \propb* outperformed \propc~on Air because the classifiers were accurate enough to make up for the lack of disagreement. This may be because Air had larger higher-order families than other data sets, so each bootstrap was less representative of the true neighborhood. However, using separate parameters for each GNN---a true ensemble approach---was the most consistently effective way to balance the trade-off between accuracy and diversity.

\section{Limitations, other related work, and future directions} \label{sec:conclusion}

\prop's empirical success has implications for multiple research areas. Most broadly, we suggest that ensembling techniques for GNNs have been underexplored. However, generalizing our findings is complicated by the fact that \prop relies on the HON to supply the neighborhood variance, which is in turn limited to representing sequential data. Our results also suggest that the task of designing informed graph representations like HONs is essential and has been previously overshadowed by downstream learning algorithms. Other research on higher-order models has made progress toward this end, but much work remains to generalize these models and the types of dependencies they can represent. One area we consider promising for future work is graph structure learning. At present, HONs are constructed in unsupervised fashion, but future work could overcome this limitation by inferring the graph structure that best solves the given task \citep{brugere2020inferring,chen2022graph}. Unfortunately, most standard data sets for GNN tasks are pre-constructed graphs, meaning that any higher-order dependencies (sequential or otherwise) have already been lost before learning begins. By integrating graph construction into the learning process and continuing to develop more powerful GNNs, we will increase our capacity to model higher-order relationships and expand the frontier of machine learning on graphs.

\bibliography{main}
\bibliographystyle{iclr2023_conference}

\clearpage
\appendix
\section*{Appendix}
\begin{table}[ht]
\caption{Summary of key notation.}
\label{tab:notation}
\centering
\begin{tabular}{ll}
\toprule
     Symbol & Description \\
     \cmidrule(l){1-2}
     $D$ & set of nodes for training or inference \\
     $D^{(i)}$ & $i^{th}$ bootstrap of relatives sampled from $D$ \\
     $D_u^{(i)}$ & relative of node $u$ sampled for the $i^{th}$ bootstrap \\
     $\mye_k$ & set of edges in $\myg_k$ \\
     $\myg_k$ & graph with max order $k$ ($k=1$ is a FON, $k>1$ is a HON) \\
     $\gnn_i(\cdot)$ & the $i^{th}$ learner (GNN) in the ensemble \\
     $\mathbf{h}_u^{(i,t)}$ & hidden representation of node $u$ at $t^{th}$ layer (timestamp) of learner $i$ \\
     $\ell $ & number of base learners in the ensemble \\
     $\mathcal{N}_k(u)$ & neighborhood of node $u$ in $\myg_k$ \\
     $P_u^k$ & distribution for sampling relatives from $\Omega_u^k$\\
     $\mathcal{S}$ & set of observed paths used to build $\myg_k$ \\ 
     $\myv_k$ & set of nodes in $\myg_k$ \\
     $w_k(u,v) $ & weight of directed edge $(u, v)$ in $\myg_k$\\
     $\hat{\mathbf{y}}_u$ & final (pooled) predicted label for node $u$ \\
     $\hat{\mathbf{y}}_u^{(i)}$ & the $i^{th}$ learner's predicted label for node $u$ \\
     $\Omega_u^k$ & relatives of node $u$ in $\myg_k$ \\
\bottomrule
\end{tabular}
\end{table}

\clearpage
\section{Proof of Theorem \ref{th:subspace}} \label{sec:proof}

\paragraph{Preliminaries} Without loss of generality, in this section we treat the neighborhood $\mathcal{N}_k(u')$ in a graph $\myg_k$ for $k \geq 1$ as only the out-neighbors of $u'$ (i.e. $\mathcal{N}_k(u') = \{v \in \myv_k, w_k(u',v) > 0\}$) and ignore in-edges. Given a node $u' \in \myv_k$, we first define the probability that a random walker over $\myg_k$ would move from $u'$ to any $v \in \mathcal{N}_k(u')$:
\begin{equation} \label{eq:pi}
    \pi_k(u' \rightarrow v) = \frac{w_k(u',v)}{outdeg_k(u')}.
\end{equation}

Given a FON $\myg_1$ and any $u \in \myv_1$, let $u' \in \Omega_u^k$ be any relative of $u$ in a HON $\myg_k$ that is constructed from the same data as $\myg_1$. Following \citet{saebi2020efficient}, we next define the Kullback-Leibler divergence of $\mathcal{N}_k(u')$ with respect to $\mathcal{N}_1(u)$:
\begin{equation} \label{eq:kld}
    D_{KL}(\mathcal{N}_k(u') \parallel \mathcal{N}_1(u)) = \sum_{v \in \mathcal{N}_1(u)} \pi_1(u \rightarrow v) log_2 \frac{\pi_1(u \rightarrow  v)}{\pi_k(u' \rightarrow v)}.
\end{equation}

Our notation throughout this section assumes that $v \in \mathcal{N}_1(u)$ and $v \in \mathcal{N}_k(u')$; however, a relative $v' \in \Omega_v^k$ often replaces $v$ in $\mathcal{N}_k(u')$. For example, in Figure \ref{fig:toy}, C$|$A is substituted for its base node C in the neighborhood of A, but the edge (A, C$|$A) still fundamentally represents a step from C to A. GrowHON \citep{krieg2020growhon} constructs $\myg_k$ such that $u' \in \myv_k$ iff the following inequality holds:

\begin{equation} \label{eq:threshold}
    D_{KL}(\mathcal{N}_k(u') \parallel \mathcal{N}_1(u)) > \frac{m}{log_2\big(1+freq(u')\big)},
\end{equation}
where $m > 1$ is length of the sequence encoded by $u'$ (see Section \ref{sec:background}) and $freq(u') \geq 1$ is the number of times $u'$ is found in all the observed paths in $\mathcal{S}$.

\newtheorem{lemma}{Lemma}
\begin{lemma} \label{lem:neighbors}
If $u \in \myv_1$ and $u' \in \Omega_u^k$, then there exists at least one node $v \in \mathcal{N}_1(u)$ such that $\pi_1(u \rightarrow v) \neq \pi_k(u' \rightarrow v)$.
\end{lemma}
\begin{proof}
If $u' \in \Omega_u^k$, then $D_{KL}(\mathcal{N}_k(u') \parallel \mathcal{N}_1(u)) > 0$. This is because $freq(u') \geq 1$, so $log_2\big(1 + freq(u')\big) \geq 1$ and the right side of Eq. \ref{eq:threshold} cannot be negative. If $D_{KL}(\mathcal{N}_k(u') \parallel \mathcal{N}_1(u)) = 0$, then the inequality in Eq. \ref{eq:threshold} cannot hold, so $u' \notin \myv_k$ and, consequently, $u' \notin \Omega_u^k$. Additionally, if $\pi_1(u \rightarrow v) = \pi_k(u' \rightarrow v)$ for all $v \in \mathcal{N}_1(u)$, then by Gibbs' inequality we have $D_{KL}(\mathcal{N}_k(u') \parallel \mathcal{N}_1(u)) = 0$. Therefore, since $u' \in \Omega_u^k$, there must exist at least one node $v \in \mathcal{N}_1(u)$ such that $\pi_1(u \rightarrow v) \neq \pi_k(u' \rightarrow v)$.
\end{proof}

We now restate and prove Theorem \ref{th:subspace}.
\newtheorem*{theoremrest}{Theorem \ref{th:subspace}}
\begin{theoremrest}
Let $\myg_1$ and $\myg_k$ be a FON and HON, respectively, both constructed from the same input $\mathcal{S}$. Let $\mathcal{N}_1(u)$ and $\mathcal{N}_k(u)$ denote the neighborhoods of any node $u$ in $\myg_1$ and $\myg_k$, respectively. Let $\textsc{\footnotesize{AGGREGATE}}(\cdot)$ represent any symmetric neighborhood aggregation function. If $u \in \myv_1$ and $u' \in \Omega_u^k$, then $\textsc{\footnotesize{AGGREGATE}}(\mathcal{N}_k(u'))$ is a biased estimator of $\textsc{\footnotesize{AGGREGATE}}(\mathcal{N}_1(u))$. 
\end{theoremrest}

\begin{proof}
Without loss of generality, we consider the case in which \textsc{\footnotesize{AGGREGATE}} operates on a sample of neighbors (i.e., GraphSAGE \citep{hamilton2017inductive}). Let $\sampv \sim \mathcal{N}_k(u')$ denote a random sample drawn from $\mathcal{N}_k(u')$. Assuming we weight the sampling distribution according to edge weights, we can write the probability of sampling any node $v$ as
\begin{equation} \label{eq:probsamp}
p(\sampv=v) = \frac{w_k(u',v)}{outdeg_k(u')} = \pi_k(u' \rightarrow v),
\end{equation}

where $w_k(u',v)$ is the weight of edge $(u', v)$ in $\myg_k$ (Section \ref{sec:background}) and $outdeg_k(u')$ is the weighted out-degree of $u'$ in $\myg_k$. If we say that each $u \in \myv_k$ is represented by a real-valued feature vector $\mathbf{x}_{u}^k = [x_{u,1}^{(k)}, x_{u,2}^{(k)}, ... ,x_{u,d}^{(k)}]$, then, for a single sample drawn from $\mathcal{N}_k(u')$, we can formulate the expected value for each vector as follows:
\begin{equation}
\mathbb{E}_{\sampv \sim \mathcal{N}_k(u')}\big[\mathbf{x}_\sampv^k \big] = p(\sampv=v)~\mathbf{x}_v^k.
\end{equation}

If we combine these feature vectors for each of the $n$ nodes in $\myv_k$, then we can represent the full expectation of $\sampv$ as a neighborhood matrix
\begin{align*}
\mathbb{E}_{\sampv \sim \mathcal{N}_k(u')}[\sampv\big]
    &=
        \begin{bmatrix}
            p(\sampv = 1)~\mathbf{x}_{1}^k \\
            p(\sampv = 2)~\mathbf{x}_{2}^k \\
           \vdots \\
           p(\sampv = n)~\mathbf{x}_{n}^k
         \end{bmatrix} \\
    &=
        \begin{bmatrix}
            p(\sampv = 1)~x_{1,1}^{(k)} & p(\sampv = 1)~x_{1,2}^{(k)} & \dots & p(\sampv = 1)~x_{1,d}^{(k)} \\
            p(\sampv = 2)~x_{2,1}^{(k)} & p(\sampv = 2)~x_{2,2}^{(k)} & \dots & p(\sampv = 2)~x_{2,d}^{(k)} \\
           \vdots & \vdots & \ddots & \vdots \\
           p(\sampv = n)~x_{n,1}^{(k)} & p(\sampv = n)~x_{n,2}^{(k)} & \dots & p(\sampv = n)~x_{n,d}^{(k)}
         \end{bmatrix}.
\end{align*}

We can simplify our notation by assuming the initial feature vectors are identity features (i.e., one-hot encodings of the node indices) defined on the base nodes, i.e., for features of nodes in $\myv_1$, we have $x_{u,j}^{(1)} = \delta_{uj}$ for all $u, j \leq d = n$, where $\delta$ is the Kronecker delta. For the features of nodes in $\myv_k$, we assume that each $u' \in \Omega_u^k$ uses the same features as its base node, i.e., $x_{u',j}^{(k)} = \delta_{uj}$. Substituting these values in the above matrix gives
\begin{align*}
\mathbb{E}_{\sampv \sim \mathcal{N}_k(u')}\big[\sampv\big]
    &=
    \begin{bmatrix}
        p(\sampv = 1) & 0 & \dots & 0 \\
        0 & p(\sampv = 2) & \dots & 0 \\
       \vdots & \vdots & \ddots & \vdots \\
       0 & 0 & \dots & p(\sampv = n)
     \end{bmatrix}.
\end{align*}

Without loss of generality, let $\textsc{\footnotesize{AGGREGATE}}$ be the feature-wise $\textsc{\footnotesize{MEAN}}$ of $N$ samples drawn from $\mathcal{N}_k(u')$. We can then represent the expectation of $\textsc{\footnotesize{AGGREGATE}}$ as

\begin{align*}
    \mathbb{E}_{\sampv \sim \mathcal{N}_k(u')}\big[\textsc{\footnotesize{AGGREGATE}}(\sampv)\big]
         &=
        \begin{bmatrix}
        \frac{1}{N} \\
        \frac{1}{N} \\
        \vdots \\
        \frac{1}{N}
     \end{bmatrix}
        \begin{bmatrix}
        Np(\sampv = 1) & 0 & \dots & 0 \\
        0 & Np(\sampv = 2) & \dots & 0 \\
       \vdots & \vdots & \ddots & \vdots \\
       0 & 0 & \dots & Np(\sampv = n)
     \end{bmatrix}
     \\
     &=
        \begin{bmatrix}
            p(\sampv = 1) \\
            p(\sampv = 2) \\
           \vdots \\
           p(\sampv = n)
         \end{bmatrix} \\
    &=
        \begin{bmatrix}
           \pi_k(u' \rightarrow 1) \\
            \pi_k(u' \rightarrow 2) \\
           \vdots \\
          \pi_k(u' \rightarrow n)
         \end{bmatrix}. & \textrm{(via Eq. \ref{eq:probsamp})}
\end{align*}

If $\sampv \sim \mathcal{N}_1(u)$, we instead have
\begin{align*}
    \mathbb{E}_{\sampv \sim \mathcal{N}_1(u)}\big[\textsc{\footnotesize{AGGREGATE}}(\sampv)\big] &=
        \begin{bmatrix}
           \pi_1(u \rightarrow 1) \\
            \pi_1(u \rightarrow 2) \\
           \vdots \\
          \pi_1(u \rightarrow n)
         \end{bmatrix}.
\end{align*}

From Lemma \ref{lem:neighbors} we know that there exists at least one $v \in \mathcal{N}_1(u)$ such that $\pi_1(u \rightarrow v) \neq \pi_k(u' \rightarrow v)$. Therefore, $\mathbb{E}_{\sampv \sim \mathcal{N}_k(u')}\big[\textsc{\footnotesize{AGGREGATE}}(\sampv)\big] \neq \mathbb{E}_{\sampv \sim \mathcal{N}_1(u)}\big[\textsc{\footnotesize{AGGREGATE}}(\sampv)\big]$, and $\textsc{\footnotesize{AGGREGATE}}(\mathcal{N}_k(u'))$ is a biased estimator of $\textsc{\footnotesize{AGGREGATE}}(\mathcal{N}_1(u))$.
\end{proof}

This result holds for other common aggregators like $\textsc{\footnotesize{SUM}}$ and $\textsc{\footnotesize{MAX}}$ for single samples. For $\textsc{\footnotesize{MAX}}$, however, the sample size matters, because the expected value for the $j^{th}$ feature is the probability that $j$ is sampled at least once. We conjecture that this is problematic for larger sample sizes in weighted graphs, and is perhaps the reason why mean-pooling outperformed max-pooling for our GraphSAGE baseline during our initial experiments (Section \ref{sec:hyperparams}).

\clearpage
\section{Data details and experimental setup} \label{sec:datasum}
\begin{itemize}
    \item [\textbf{Air:}] Flight trajectories of passenger itineraries in the United States. Nodes represent airport locations (cities) and edges represent passengers flying between locations. The itineraries were retrieved from the Airline Origin and Destination Survey (DB1B) database, which is publicly available through the U.S. Bureau of Transportation Statistics\footnote{\url{https://transtats.bts.gov/}. Accessed April 14, 2022.}. We downloaded all records between Jan. 1 and Dec. 31, 2019 from the DB1BCoupon table, joined and sorted itineraries using the ``ITIN\_ID'' and ``SEQ\_NUM'' columns, and discarded any itineraries with missing origin or destination information. Node classes represent the geographical location of each airport, aggregated by standard federal region of the United States. While we extracted this particular set of paths, this database has been utilized in prior work on HONs \citep{rosvall2014memory,scholtes2017network}.
    \item [\textbf{T2D:}] Disease trajectories for type 2 diabetes patients in the state of Indiana \citep{krieg2020higher}. Nodes represent ICD9 diagnosis codes and edges represent sequential diagnoses. Following prior work, we only preserved the first occurrence of each diagnosis code for each patient; however, we did not split trajectories based on the period of time between diagnoses. Node classes are the chapters of each ICD9 code, which represent categories of disease classification.
    \item [\textbf{Wiki:}] Clickstreams of users playing the Wikispeedia game, in which a player attempts to navigate from a source to a target article by clicking only Wikipedia links \citep{west2009wikispeedia}. Here, nodes represent Wikipedia articles, and edges represent user clicks between articles. We included both finished and unfinished paths. Node classes are the subjects of each article.
    \item [\textbf{Mag:}] Readership trajectories for a large online magazine from Jan 1. to Apr. 15, 2020 \citep{wang2020calendar}. Nodes represent online content (articles, games, etc.), and edges represent sequential clicks by users. We only considered trajectories in which the user visited at least three nodes during a session. Node classes represent the content type (magazine, culture, news, or humor).
    \item [\textbf{Mag+:}] An expanded version of Mag which also includes data from July 1, 2019 through Dec. 31, 2019, as well as two additional content types (books, home). Due to its size and similarity to Mag, we excluded it from link prediction experiments.
    \item [\textbf{Ship:}] Trajectories of global shipping activity \citep{saebi2020higher}. Nodes represent ports, and edges represent ships traveling between ports. Node classes are provinces, as defined by the Marine Ecoregions of the World (MEOW) system for classifying oceans and waterways \citep{spalding2007marine}.
\end{itemize}

For node classification, we used stratified 5-fold cross validation \citep{shchur2018pitfalls} and reported the mean micro F1-score. For link prediction, we generated training and testing sets by randomly sampling 10\% of positive node pairs and an equal number of negative node pairs. In addition to hiding all edges (in both directions) between testing pairs, for experiments using $\myg_2$ we also hid all edges between any pair of nodes in the same higher-order families as each testing pair. We repeated each experiment five times and reported the mean area under the precision-recall curve (AUPRC) \citep{yang2015evaluating}. All experiments utilized the same training and testing splits. For models trained on $\myg_1$ we used identity features (i.e., one-hot encodings of the node indices) for each node. For models trained on $\myg_2$, we used the same features as $\myg_1$ such that each $u' \in \Omega_u^2$ shared the same features as its base node $u$.

\clearpage
\section{Model tuning and hyperparameter setup} \label{sec:hyperparams}

We manually tuned hyperparameters for each model. Table \ref{tab:hyperparameters} summarizes the configurations that we evaluated in reporting the main results in Tables \ref{tab:nc} and \ref{tab:lp}. We used the Adam optimizer with a learning rate of 0.0005 for GAT and GATv2, and a learning rate of 0.01 for all other models. For minibatch models (GraphSAGE, GIN, and \prop), we tested batch sizes of 16, 32, and 64. Unless otherwise noted, all results reported in Tables \ref{tab:nc}, \ref{tab:lp}, \ref{tab:ncfull}, and \ref{tab:lpfull} were the best-performing configuration for each model as averaged across all testing folds.

In general, the most impactful hyperparameter was the number of GNN layers. As discussed in Section \ref{sec:modeldepth}, on Air, T2D, and Mag, all baselines performed best with only a single layer for both node classification and link prediction. On Wiki, several models performed best with 3 layers (Figure \ref{fig:depthloss}). GCNII, which uses residual connections and is thus designed with deep GNNs in mind \citep{chen2020simple}, was the only model we tested with more than 3 layers. For most models, increasing the number of hidden units per layer above 256 had little to no effect on performance. The exception to this was \propb* on Air, which performed best with 2,048 hidden units (Section \ref{sec:ensdiv}).

Other hyperparameters varied by data set. The number of neighbors sampled ($\vert\mathcal{N}\vert$) was very important on the dense graphs (T2D and Mag). In all cases, we found that increasing the sample size of the first layer was the most important, so we fixed the sample size at the second layer to 1 (per first-layer sample). Baselines preferred 512 samples for T2D and 256 for Mag. On the other data sets, 64 neighbor samples was sufficient for all models. \prop preferred 256 samples per base learner (the max that we tested) for T2D and 128 for Mag. The same trend held for GraphSAINT's subgraph sampling, which preferred more roots on graphs that were dense and had more nodes (512 for Air and Wiki, 1024 for T2D, and 2048 for Mag, 4096 for Mag+, and 2048 for Ship). On all data sets, PathGCN performed best with at least 100 random walk samples per node.

For neighborhood aggregators, max-pooling was in all cases inferior to mean-pooling and sum-pooling, likely due to the inability of max-pooling to appropriately distinguish dense neighborhoods \citep{xu2018how}.

\begin{table}[ht]
\caption{Hyperparameters evaluated for each model.}
\label{tab:hyperparameters}
\centering
\begin{tabular}{lcc}
\toprule
\multicolumn{3}{c}{Model-specific hyperparameters} \\
\cmidrule(l{11.5em}r{11.5em}){1-3}
Model & \# Layers & Other \\
\cmidrule(lr){1-3}
GCN & $\{1,2,3\}$ & --- \\
GCNII & $\{1,2,3,4,6,8\}$ & $\alpha=0.5, \lambda=1.0$\\
GAT & $\{1,2,3\}$ & attnheads $=\{4,8,16\}$ \\
GATv2 & $\{1,2,3\}$ &  attnheads $=\{4,8,16\}$ \\
GraphSAGE & $\{1,2,3\}$ & $|\mathcal{N}| = \{32,64,128,256,512,1024\} \times \{1,2,4,8 \} \times \{1\}$ \\
& & $\textsc{\footnotesize{AGGREGATE}}$ = \{$\textsc{\footnotesize{MEANPOOL}}$, $\textsc{\footnotesize{MAXPOOL}}$\} \\
GIN & $\{1,2,3\}$ & $|\mathcal{N}| = \{32,64,128,256,512,1024 \} \times \{1,2,4,8\} \times \{1\}$\\
GraphSAINT & $\{1,2,3\}$ & nroots $=\{256,512,1024,2048, 4096\}$, walklen $=\{2,3,4\}$ \\
PathGCN & $\{1,2,3\}$ & nwalks $=\{10,50,100,200\}$, walklen = $\{2,3,4,5\}$ \\
HO-GNN & $\{1,2,3\}$ & --- \\ 
SEAL & $\{1,2,3\}$ & hops $=\{1,2,3\}$ \\ 
DGE (all) & $\{1,2,3\} $ & $\ell = \{4,8,16\}$, $|\mathcal{N}| = \{32,64,128,256\} \times \{1,2,4\} \times \{1\}$ \\
\cmidrule(lr){1-3}
\multicolumn{3}{c}{General hyperparameters} \\
\cmidrule(l{13em}r{13em}){1-3}
Model & Dropout & Hidden units (per layer) \\
\cmidrule(lr){1-3}
All & 0.4 & \{128, 256, 512, 1024, 2048\}  \\
\bottomrule
\end{tabular}
\end{table}

\clearpage
\section{Additional pooling mechanisms} \label{sec:additionalpooling}
In addition to Eqs. \ref{eq:propa}, \ref{eq:propb}, and \ref{eq:propc}, we evaluated
an attention mechanism (inspired by GAT \citep{velivckovic2018graph}) which used self-attention to pool the outputs of all sampled relatives. 
We computed attention coefficients $e_{ij}$ for each $i, j \leq \ell$ according to
\begin{equation}
e_{ij} = F\Big(\mathbf{W}\big(\gnn_i\ (D^{(i)}_u)\big), \mathbf{W}\big(\gnn_j\ (D^{(j)}_u)\big)\Big),
\end{equation}
where $F$ is a single-layer feed-forward network with a LeakyReLU activation, $\mathbf{W} \in \mathbb{R}^{2d \times d}$ is a trainable weight matrix, and $\gnn_i(u) = \mathbf{h}_{u}^{(i,t)}$ (as in Eqs. \ref{eq:propa} and \ref{eq:propb}). We then computed the final
attention coefficients $\alpha_{ij}$ according to
\begin{equation}
\alpha_{ij} = \text{softmax}_{i}(e_{ij}) = \frac{\exp(e_{ij})}{\sum_{m = 1}^{\ell} \exp(e_{im})}.
\end{equation}

We repeated this procedure for $K$ separate attention heads. Whereas the original GAT layer averages
the outputs of the attention heads before applying a non-linear activation \citep{velivckovic2018graph},
we found that passing the same output to another feed-forward network provided better results.
Formally, letting $i$ be the index of we compute the vector of class probabilities $\mathbf{\hat y}_u$ by
\begin{equation}
\mathbf{\hat y}_u = \sigma\Biggl(F'\Big(\sigma'\Big(\frac{1}{K}\sum_{n=1}^K \underset{j = 1}{\overset{\ell}{\Big\Vert}} \sum_{i = 1}^{\ell} \alpha^{(n)}_{ij} \mathbf{W}^{(n)} \gnn_j\ (D_u^{(j)}) \Big) \Big) \Biggl),
\end{equation}
where $\Vert$ is vector concatenation, $F'$ is a feed-forward network with output dimension $d/4$ (followed by batch normalization and a LeakyReLU activation), $\alpha_{ij}^{(n)}$ and $\mathbf{W}^{(n)}$ represent the attention coefficients and linear transformation, respectively, for the $n^{th}$ attention head, and $\sigma'$ is a LeakyReLU activation. For our experiments, we set $K=5$, $d=128$, and a negative slope coefficient of 0.3 for all LeakyReLU activations. We refer to this model as DGE-attn and discuss its performance in Appendix \ref{sec:supresults}.

\clearpage
\section{Additional experimental results} \label{sec:supresults}

To support our claim that existing GNNs cannot effectively use $\myg_2$ as input, we also tested baselines using $\myg_2$ as the input graph for Air, T2D, Wiki, and Mag. As we expected, because each neighborhood in $\myg_2$ is only a subspace of its neighborhood in $\myg_1$, performance was generally worse (or about the same) when compared to $\myg_1$. Tables \ref{tab:ncfull} and \ref{tab:lpfull} detail the full results. 

Table \ref{tab:ncfull} also includes results for a GNN ensemble (\propc) trained on $\myg_1$, i.e. without relative sampling and pooling. Overall, we found that the performance was slightly better than GraphSAGE (the base GNN), but in all cases fell short of \propc~with relative sampling and pooling. The final result included in Table \ref{tab:ncfull} is for one additional pooling mechanism, DGE-attn, which pooled representations from the sampled relatives via a self-attention mechanism. Since it did not generalize particularly well in the node classification experiments, we did not discuss it in the main manuscript and did not evaluate its performance on link prediction. More details are available in Appendix \ref{sec:additionalpooling}. Table \ref{tab:lpfull} also includes the full link prediction results for baselines, which we excluded from Table \ref{tab:lp} because of space constraints in the main manuscript. 

\begin{figure}[ht]
    \centering
    \includegraphics[width=0.4\linewidth]{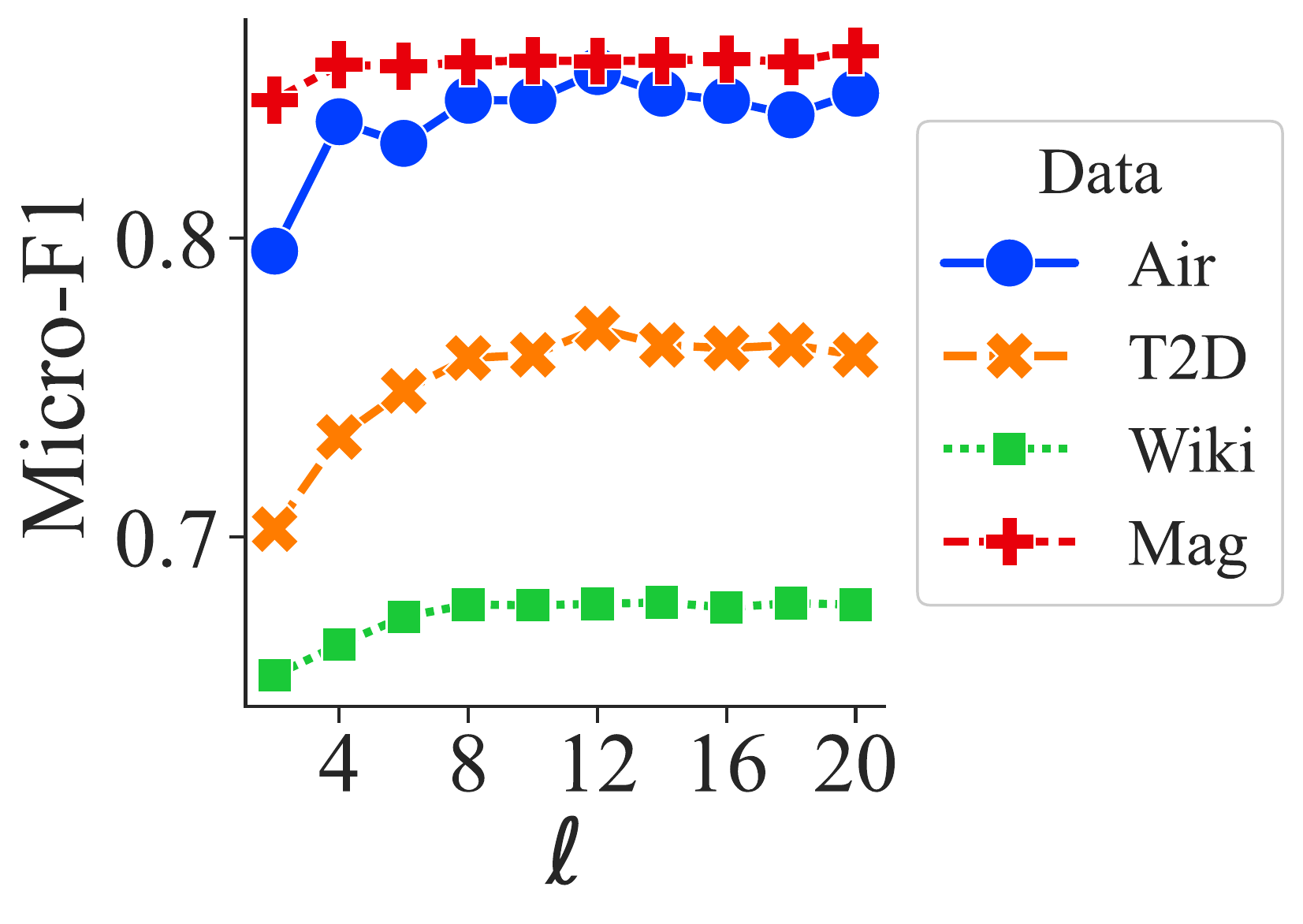}
    \caption{Node classification performance for \propc~as a function of the number of base learners ($\ell$).
    }
    \label{fig:enssize}
\end{figure}

\begin{figure}[t]
    \centering
    \begin{subfigure}{\linewidth}
    \includegraphics[width=\linewidth]{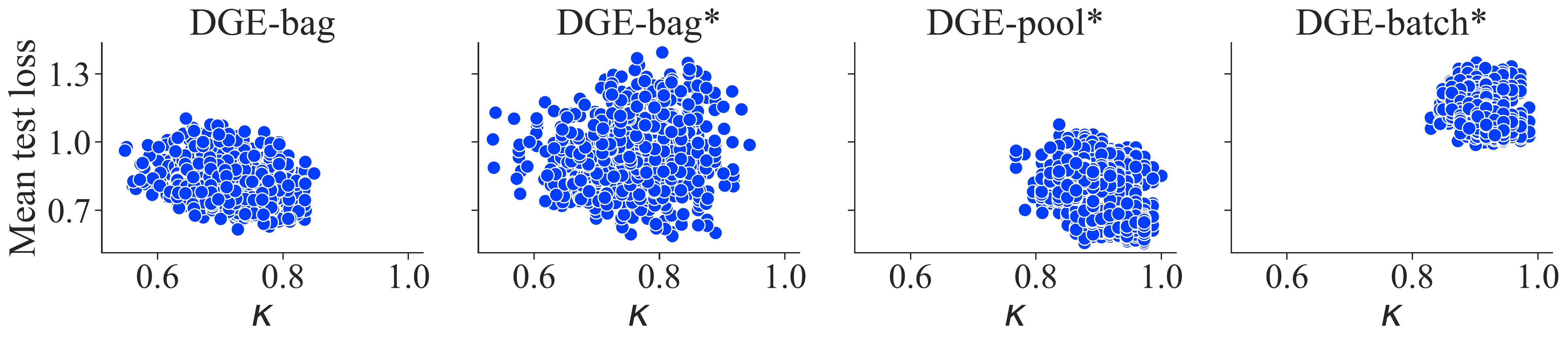}
    \caption{Air results.}
    \end{subfigure}
    \par\bigskip
    \begin{subfigure}{\linewidth}
    \includegraphics[width=\linewidth]{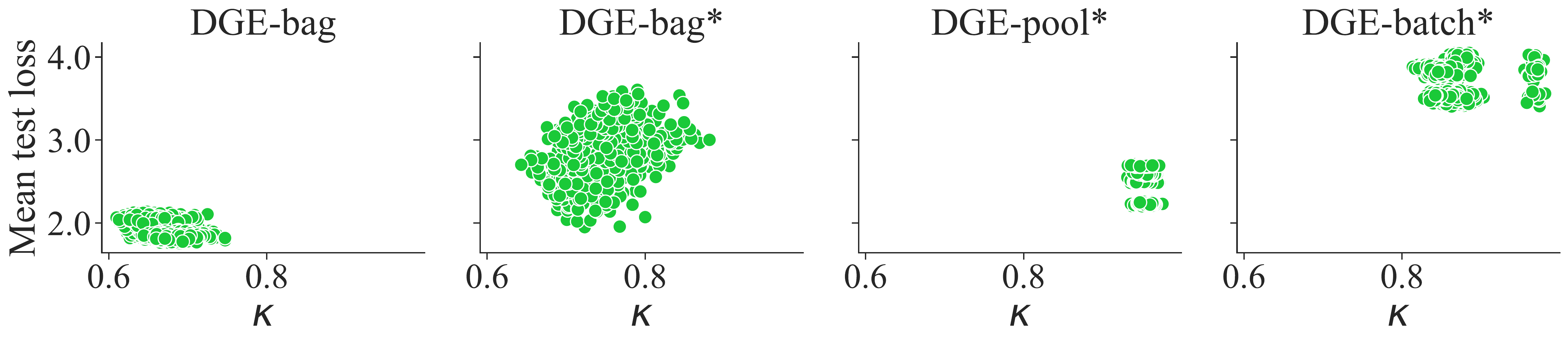}
    \caption{Wiki results.}
    \end{subfigure}
    \par\bigskip
    \begin{subfigure}{\linewidth}
    \includegraphics[width=\linewidth]{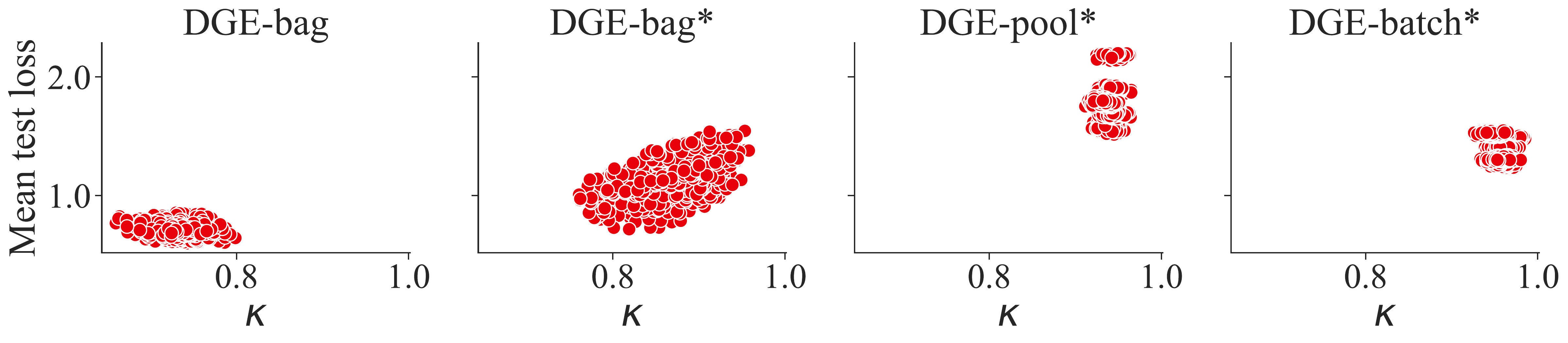}
    \caption{Mag results.}
    \end{subfigure}
    \caption{Mean node classification loss of each pair of classifiers, plotted as a function of Cohen's kappa (lower values indicate lower agreement). Each point represents one pairwise comparison between the 16 classifiers in each of the 5 testing folds. All plots contain the same number of points.}
    \label{fig:kappassup}
\end{figure}

Figure \ref{fig:enssize} shows \propc's performance as a function of the number of base learners ($\ell$) on four data sets. We found that 8-12 base learners produced consistently strong results. Figure \ref{fig:kappassup} shows classifier accuracy and diversity (Section \ref{sec:ensdiv}) for the T2D, Wiki, and Mag data sets, as a supplement to Figure \ref{fig:kappas}.

\begin{table}[ht]
\caption{Additional node classification results (micro F1) for baselines using $\myg_2$ as input. Bold font indicates the best result for each data set.}
\label{tab:ncfull}
\centering
\begin{tabular}{lccccc}
\toprule
Model & Input & Air & T2D & Wiki & Mag \\
\cmidrule(lr){1-6}
GCN         & $\myg_1$ & 0.818 \scriptsize $\pm$ 0.03 & 0.480 \scriptsize $\pm$ 0.02 & 0.643 \scriptsize $\pm$ 0.01 & 0.796 \scriptsize $\pm$ 0.01 \\
            & $\myg_2$ & 0.805 \scriptsize $\pm$ 0.05 & 0.500 \scriptsize $\pm$ 0.02 & 0.665 \scriptsize $\pm$ 0.02 & 0.744 \scriptsize $\pm$ 0.02 \\
GCNII       & $\myg_1$ & 0.845 \scriptsize $\pm$ 0.05 & 0.511 \scriptsize $\pm$ 0.02 & 0.654 \scriptsize $\pm$ 0.02 & 0.801 \scriptsize $\pm$ 0.01 \\
            & $\myg_2$ & 0.833 \scriptsize $\pm$ 0.05 & 0.537 \scriptsize $\pm$ 0.03 & 0.657 \scriptsize $\pm$ 0.02 & 0.816 \scriptsize $\pm$ 0.01 \\
GAT         & $\myg_1$ & 0.804 \scriptsize $\pm$ 0.03 & 0.282 \scriptsize $\pm$ 0.10 & 0.639 \scriptsize $\pm$ 0.02 & 0.487 \scriptsize $\pm$ 0.06 \\
            & $\myg_2$ & 0.817 \scriptsize $\pm$ 0.05 & 0.253 \scriptsize $\pm$ 0.03 & 0.646 \scriptsize $\pm$ 0.03 & 0.512 \scriptsize $\pm$ 0.04\\
GATv2       & $\myg_1$ & 0.838 \scriptsize $\pm$ 0.03 & 0.292 \scriptsize $\pm$ 0.07 & 0.643 \scriptsize $\pm$ 0.03 & 0.495 \scriptsize $\pm$ 0.05 \\
            & $\myg_2$ & 0.842 \scriptsize $\pm$ 0.05 & 0.241 \scriptsize $\pm$ 0.10 & 0.646 \scriptsize $\pm$ 0.03 & 0.512 \scriptsize $\pm$ 0.13 \\
GraphSAGE   & $\myg_1$ & 0.781 \scriptsize $\pm$ 0.04  & 0.654 \scriptsize $\pm$ 0.04 & 0.625 \scriptsize $\pm$ 0.02 & 0.808 \scriptsize $\pm$ 0.02 \\
            & $\myg_2$ & 0.688 \scriptsize $\pm$ 0.02 & 0.516 \scriptsize $\pm$ 0.01 & 0.633 \scriptsize $\pm$ 0.01 & 0.818 \scriptsize $\pm$ 0.01 \\
GIN         & $\myg_1$ & 0.745 \scriptsize $\pm$ 0.02 & 0.673 \scriptsize $\pm$ 0.04 & 0.636 \scriptsize $\pm$ 0.02 & 0.826 \scriptsize $\pm$ 0.02 \\
            & $\myg_2$ & 0.660 \scriptsize $\pm$ 0.02 & 0.637 \scriptsize $\pm$ 0.03 & 0.625 \scriptsize $\pm$ 0.02 & 0.812 \scriptsize $\pm$ 0.02 \\
GraphSAINT  & $\myg_1$ & 0.802 \scriptsize $\pm$ 0.02 & 0.600 \scriptsize $\pm$ 0.07 & 0.664 \scriptsize $\pm$ 0.01 & 0.821 \scriptsize $\pm$ 0.02 \\
            & $\myg_2$ & 0.710 \scriptsize $\pm$ 0.03 & 0.567 \scriptsize $\pm$ 0.03 & 0.661 \scriptsize $\pm$ 0.02 & 0.828 \scriptsize $\pm$ 0.01 \\
HONEM       & $\myg_1,\myg_2$  & 0.805 \scriptsize $\pm$ 0.04  & 0.566 \scriptsize $\pm$ 0.02  & 0.588 \scriptsize $\pm$ 0.01  & 0.728 \scriptsize $\pm$ 0.02  \\
\cmidrule(lr){1-6}
\propc & $\myg_1$ & 0.772 \scriptsize $\pm$ 0.05 & 0.694 \scriptsize $\pm$ 0.04 & 0.645 \scriptsize $\pm$ 0.02 & 0.831 \scriptsize $\pm$ 0.01 \\
\cmidrule(lr){1-6}
\propa & $\myg_2$ & 0.825 \scriptsize $\pm$ 0.04 & 0.501 \scriptsize $\pm$ 0.06 & 0.615 \scriptsize $\pm$ 0.02 & 0.790 \scriptsize $\pm$ 0.02 \\
\propa* & $\myg_2$ & 0.810 \scriptsize $\pm$ 0.04 & 0.439 \scriptsize $\pm$ 0.03 & 0.577 \scriptsize $\pm$ 0.02 & 0.761 \scriptsize $\pm$ 0.02 \\
\propb & $\myg_2$ & 0.839 \scriptsize $\pm$ 0.03 & 0.735 \scriptsize $\pm$ 0.03 & 0.671 \scriptsize $\pm$ 0.01 & 0.860 \scriptsize $\pm$ 0.01 \\
\propb* & $\myg_2$ & \textbf{0.865} \scriptsize $\pm$ 0.02 & 0.555 \scriptsize $\pm$ 0.07 & 0.599 \scriptsize $\pm$ 0.04 & 0.775 \scriptsize $\pm$ 0.01 \\
\propc & $\myg_2$ & 0.856 \scriptsize $\pm$ 0.02 & \textbf{0.770} \scriptsize $\pm$ 0.04 & \textbf{0.681} \scriptsize $\pm$ 0.00 & \textbf{0.871} \scriptsize $\pm$ 0.01 \\
\propc* & $\myg_2$ & 0.766 \scriptsize $\pm$ 0.04 & 0.719 \scriptsize $\pm$ 0.04 & 0.644 \scriptsize $\pm$ 0.02 & 0.841 \scriptsize $\pm$ 0.02 \\
\propd* & $\myg_2$ & 0.764 \scriptsize $\pm$ 0.03 & 0.646 \scriptsize $\pm$ 0.01 & 0.623 \scriptsize $\pm$ 0.01 & 0.818 \scriptsize $\pm$ 0.01 \\
\text{DGE-attn} & $\myg_2$ & 0.829 \scriptsize $\pm$ 0.03 & 0.549 \scriptsize $\pm$ 0.05 & 0.618 \scriptsize $\pm$ 0.02 & 0.775 \scriptsize $\pm$  0.05 \\
\bottomrule
\scriptsize{* shared parameters}
\end{tabular}
\end{table}

\begin{table}[ht]
\caption{Full link prediction results (AUPRC) for baselines. Bold font indicates the best result for each data set.}
\label{tab:lpfull}
\centering
\begin{tabular}{lcccccc}
\toprule
Model & Input & Air & T2D & Wiki & Mag & Ship \\
\cmidrule(lr){1-7}
GCN         & $\myg_1$ & 0.794 \scriptsize $\pm$ 0.01 & 0.793 \scriptsize $\pm$ 0.00 & 0.741 \scriptsize $\pm$ 0.01 & 0.742 \scriptsize $\pm$ 0.00 & 0.849 \scriptsize $\pm$ 0.00 \\
            & $\myg_2$ & 0.777 \scriptsize $\pm$ 0.02 & 0.781 \scriptsize $\pm$ 0.00 & 0.749 \scriptsize $\pm$ 0.01 & 0.753 \scriptsize $\pm$ 0.00 & --- \\
GCNII       & $\myg_1$ & 0.806 \scriptsize $\pm$ 0.01 & 0.754 \scriptsize $\pm$ 0.01 & 0.781 \scriptsize $\pm$ 0.01 & 0.750 \scriptsize $\pm$ 0.00 & 0.848 \scriptsize $\pm$ 0.01 \\
            & $\myg_2$ &  0.771 \scriptsize $\pm$ 0.01 & 0.732 \scriptsize $\pm$ 0.00 & 0.790 \scriptsize $\pm$ 0.01 & 0.759 \scriptsize $\pm$ 0.00 & --- \\
GAT         & $\myg_1$  & 0.786 \scriptsize $\pm$ 0.01 & 0.698 \scriptsize $\pm$ 0.02 & 0.794 \scriptsize $\pm$ 0.01 & 0.644 \scriptsize $\pm$ 0.00 & 0.764 \scriptsize $\pm$ 0.00 \\
            & $\myg_2$ & 0.768 \scriptsize $\pm$ 0.01 & 0.681 \scriptsize $\pm$ 0.01 & 0.801 \scriptsize $\pm$ 0.01 & 0.641 \scriptsize $\pm$ 0.00 & --- \\
GATv2       & $\myg_1$ & 0.771 \scriptsize $\pm$ 0.01 & 0.701 \scriptsize $\pm$ 0.01 & 0.797 \scriptsize $\pm$ 0.01 & 0.645 \scriptsize $\pm$ 0.01 & 0.797  \scriptsize $\pm$ 0.00 \\
            & $\myg_2$ & 0.784 \scriptsize $\pm$ 0.01 & 0.684 \scriptsize $\pm$ 0.00 & 0.796 \scriptsize $\pm$ 0.01 & 0.660 \scriptsize $\pm$ 0.01  & --- \\
GraphSAGE   & $\myg_1$  & 0.782 \scriptsize $\pm$ 0.02 & 0.713 \scriptsize $\pm$ 0.01 & 0.758 \scriptsize $\pm$ 0.01 & 0.717 \scriptsize $\pm$ 0.00 & 0.820 \scriptsize $\pm$ 0.00 \\
            & $\myg_2$ & 0.757 \scriptsize $\pm$ 0.02 &  0.712 \scriptsize $\pm$ 0.01 & 0.756 \scriptsize $\pm$ 0.01 & 0.716 \scriptsize $\pm$ 0.00 & --- \\
GIN         & $\myg_1$  & 0.800 \scriptsize $\pm$ 0.01 & 0.704 \scriptsize $\pm$ 0.00 & 0.745 \scriptsize $\pm$ 0.01 & 0.717 \scriptsize $\pm$ 0.00 & 0.782 \scriptsize $\pm$ 0.00 \\
            & $\myg_2$ & 0.748 \scriptsize $\pm$ 0.02 & 0.702 \scriptsize $\pm$ 0.00 & 0.743 \scriptsize $\pm$ 0.01 & 0.716 \scriptsize $\pm$ 0.00 & --- \\
GraphSAINT  & $\myg_1$ & 0.718 \scriptsize $\pm$ 0.01 & 0.797 \scriptsize $\pm$ 0.00 &0.688 \scriptsize $\pm$ 0.01 & 0.750 \scriptsize $\pm$ 0.00 & 0.862 \scriptsize $\pm$ 0.01 \\
            & $\myg_2$  & 0.702 \scriptsize $\pm$ 0.02 & 0.791 \scriptsize $\pm$ 0.01 & 0.694 \scriptsize $\pm$ 0.01 & 0.756 \scriptsize $\pm$ 0.00 & --- \\
SEAL & $\myg_1$ & 0.818 \scriptsize $\pm$ 0.02 & 0.754 \scriptsize $\pm$ 0.01 & 0.834 \scriptsize $\pm$ 0.01 & 0.751 \scriptsize $\pm$ 0.00 & 0.887 \scriptsize $\pm$ 0.01 \\
HONEM       & $\myg_1,\myg_2$  & 0.697 \scriptsize $\pm$ 0.02 & 0.818 \scriptsize $\pm$ 0.00 & 0.589 \scriptsize $\pm$ 0.01 & 0.769 \scriptsize $\pm$ 0.00 & 0.815 \scriptsize $\pm$ 0.00 \\
HO-GNN & $\myg_2$ & 0.811 \scriptsize $\pm$ 0.00 & 0.789 \scriptsize $\pm$ 0.00 & 0.820 \scriptsize $\pm$ 0.02 & 0.879 \scriptsize $\pm$ 0.00 & 0.856 \scriptsize $\pm$ 0.00 \\
\cmidrule(lr){1-7}
\propa & $\myg_2$ & \textbf{0.886} \scriptsize $\pm$ 0.01 & 0.815 \scriptsize $\pm$ 0.01 & 0.774 \scriptsize $\pm$ 0.01 & 0.802 \scriptsize $\pm$ 0.00 & \textbf{0.910} \scriptsize $\pm$ 0.00 \\
\propa* & $\myg_2$ & 0.856 \scriptsize $\pm$ 0.01 & 0.779 \scriptsize $\pm$ 0.01 & 0.712 \scriptsize $\pm$ 0.02 & 0.898 \scriptsize $\pm$ 0.00 & 0.904 \scriptsize $\pm$ 0.00 \\
\propb & $\myg_2$ & 0.851 \scriptsize $\pm$ 0.02 & \textbf{0.920} \scriptsize $\pm$ 0.00 & 0.838 \scriptsize $\pm$ 0.03 & 0.913 \scriptsize $\pm$ 0.00 & 0.898 \scriptsize $\pm$ 0.00 \\
\propb* & $\myg_2$ & 0.845 \scriptsize $\pm$ 0.01 & 0.901 \scriptsize $\pm$ 0.00 & 0.802 \scriptsize $\pm$ 0.06 & 0.769 \scriptsize $\pm$ 0.01 & 0.815 \scriptsize $\pm$ 0.00 \\
\propc & $\myg_2$ & \textbf{0.887} \scriptsize $\pm$ 0.00 & 0.907 \scriptsize $\pm$ 0.00 & \textbf{0.876} \scriptsize $\pm$ 0.01 & \textbf{0.921} \scriptsize $\pm$ 0.00  & 0.895 \scriptsize $\pm$ 0.00 \\
\propc* & $\myg_2$ & 0.862 \scriptsize $\pm$ 0.02 & 0.894 \scriptsize $\pm$ 0.00 & 0.856 \scriptsize $\pm$ 0.00 & 0.891 \scriptsize $\pm$ 0.00 & 0.891 \scriptsize $\pm$ 0.00 \\
\propd* & $\myg_2$ & 0.853 \scriptsize $\pm$ 0.03 & 0.871 \scriptsize $\pm$ 0.01 & 0.830 \scriptsize $\pm$ 0.01 & 0.865 \scriptsize $\pm$ 0.00 & 0.865 \scriptsize $\pm$ 0.00 \\
\bottomrule
\scriptsize{* shared parameters}
\end{tabular}
\end{table}

\begin{table}[t]
\caption{Node classification results (micro F1) under various parameter budgets. Bold font indicates the best result for each budget and data set.}
\label{tab:modelwidth2}
\centering
\begin{tabular}{lccccccccc}
\toprule
& & \multicolumn{4}{c}{Total parameters (Wiki)} & \multicolumn{4}{c}{Total parameters (Mag)} \\
Model & Layers & 50k & 100k & 500k & 1m & 50k & 100k & 500k & 1m \\
\cmidrule(lr){1-2} \cmidrule(lr){3-6} \cmidrule(lr){7-10}
GCNII  & 2 & 0.60 & 0.60 & 0.60 & 0.60 & 0.78 & 0.79 & 0.79 & 0.79 \\
                        & 4 & 0.58 & 0.61 & 0.62 & 0.63 & 0.79 & 0.80 & 0.80 & 0.80 \\
                        & 8 & 0.60 & 0.61 & 0.62 & 0.62 & 0.77 & 0.79 & 0.79 & 0.79 \\
GATv2  & 2 & 0.46 & 0.52 & 0.61 & 0.62 & 0.26 & 0.30 & 0.37 & 0.40 \\
                        & 4 & 0.40 & 0.47 & 0.51 & 0.55 & --- & --- & --- & --- \\
GIN    & 2 & 0.59 & 0.61 & 0.62 & 0.62 & 0.82 & 0.82 & 0.83 & 0.83 \\
                        & 4 & 0.50 & 0.54 & 0.61 & 0.62 & 0.66 & 0.76 & 0.79 & 0.79 \\
GraphSAINT & 2 & 0.61 & \textbf{0.64} & 0.66 & 0.66 & 0.79 & 0.79 & 0.81 & 0.82 \\
                        & 4 & 0.55 & 0.62 & 0.66 & 0.66 & 0.54 & 0.65 & 0.70 & 0.74 \\
\cmidrule(lr){1-2} \cmidrule(lr){3-6} \cmidrule(lr){7-10}
DGE-bag                 & 2 & 0.52 & 0.58 & \textbf{0.68} & \textbf{0.69} & 0.81 & \textbf{0.84} & \textbf{0.86} & \textbf{0.87} \\
DGE-bag*                & 2 & \textbf{0.63} & \textbf{0.64} & 0.65 & 0.65 & \textbf{0.83} & \textbf{0.84} & 0.85 & 0.85 \\
\bottomrule
\scriptsize{* shared parameters}
\end{tabular}
\end{table}

\clearpage
\section{Training time and convergence} \label{sec:traintime}
\begin{figure}[th]
\begin{subfigure}{\textwidth}
    \centering
    \includegraphics[width=\linewidth]{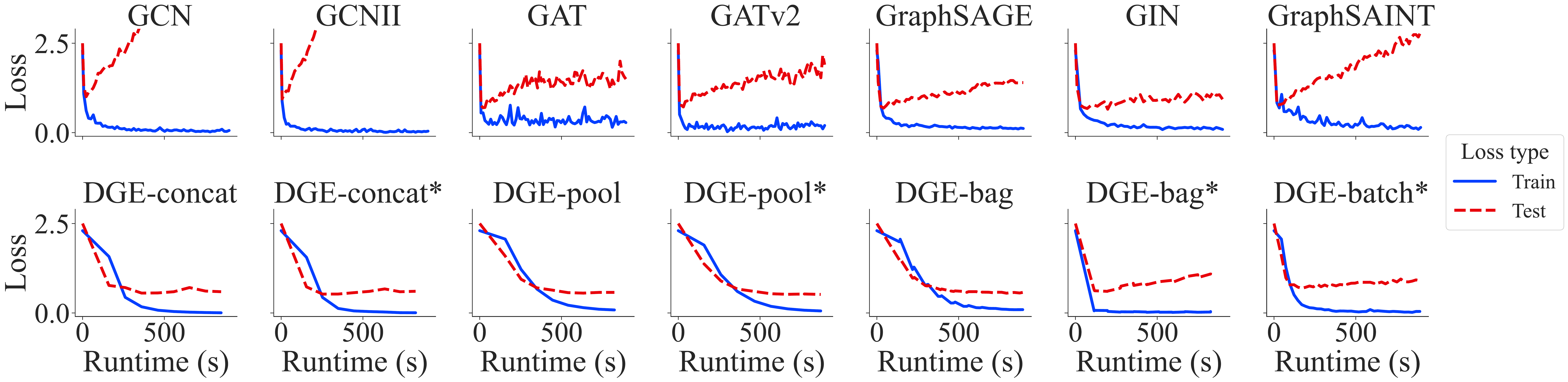}
    \caption{Air results.}
    \label{fig:timelossair}
\end{subfigure}
\par\bigskip
\begin{subfigure}{\textwidth}
    \centering
    \includegraphics[width=\linewidth]{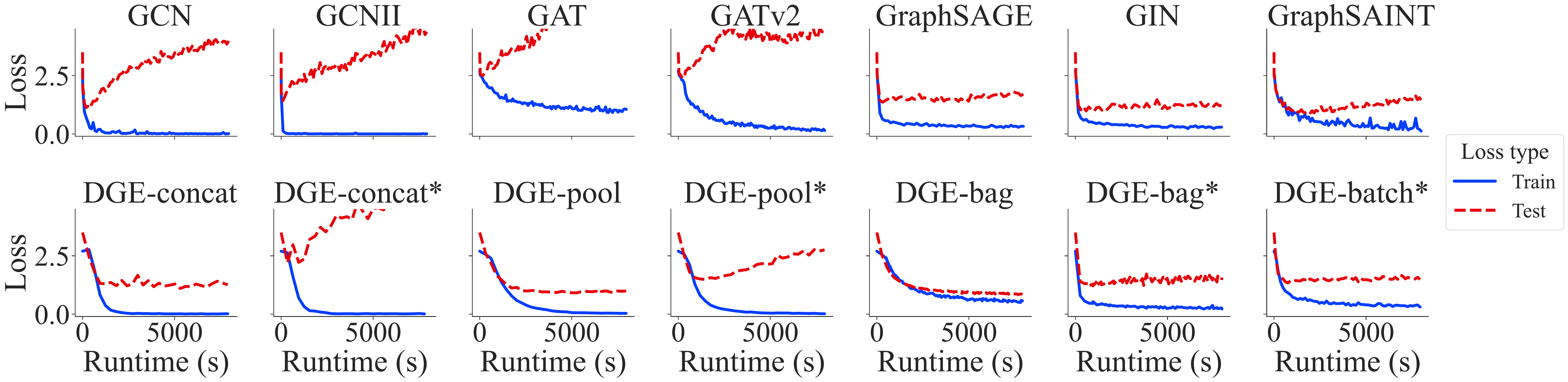}
    \caption{T2D results.}
    \label{fig:timelosst2d}
\end{subfigure}
\caption{Node classification loss for 2-layer models on the first testing fold, shown as a function of training time. Runtime is measured as wallclock time from the start of the first epoch. High loss values are truncated to preserve scale. Similar trends held for all testing folds.}
\label{fig:timeloss}
\end{figure}

We analyzed the additional time cost incurred by \prop by plotting training and testing loss as a function of total runtime for Air and T2D. We used an NVIDIA GeForce GTX TITAN X GPU and dual 8-core 2.4GHz Intel Xeon processors for each model. As Figure \ref{fig:timeloss} shows, \prop converged more slowy but generalized significantly better than all baselines. While the baselines typically converged quickly on the training set, they also rapidly began to overfit on the testing set. The increase in test error was sharper for the full-batch models like GCN and GCNII since they completed more epochs in a shorter period of time. \propc~, and, to a lesser extent, \propb, generalized extremely well. These observations reflect overall model performance as reported in Table \ref{tab:nc}. One difference is that many of the baseline results reported in Table \ref{tab:nc} used only a single GNN layer (Section \ref{sec:modeldepth}). For the results in Figure \ref{fig:timeloss}, we fixed the number of GNN layers to 2 in order to confirm that the poor generalization of baselines (Section \ref{sec:modeldepth}) was not due to lack of model capacity or underfitting the training set. This is why some models, like GCNII on Air, were relatively competitive in the main results but show high generalization error here. 

The sample-based methods, including \prop, incurred significant overhead from neighbor sampling (over 95\% of total runtime). This was exacerbated by the fact that all models preferred a high number of neighbor samples at the first layer, so repeatedly sampling a single neighbor at the second layer (Appendix \ref{sec:hyperparams}) produced substantial procedural overhead. \prop would train significantly faster with a more efficient mechanism for sampling these two-hop neighbors (such as pre-fetching).

All variants of \prop converged in fewer epochs than baselines, likely because \prop sees several subgraphs for each node during the same epoch (by virtue of being an ensemble) and because $\myg_k$ is sparser than $\myg_1$. On T2D, convergence took approximately 100 epochs for \prop, 100-150 for minibatch baselines, and 1000-1500 for full-batch baselines. On other data sets, \prop generally converged in approximately 15-20 epochs, minibatch baselines converged in 20-30 epochs, and other full-batch baselines took 100-200 epochs (except Mag and Mag+, which took the full-batch models 500+ epochs). 

\clearpage
\section{Analysis of node characteristics} \label{sec:nodechar}
\begin{figure}[ht]
\begin{subfigure}{\linewidth}
    \centering
    \includegraphics[width=\linewidth]{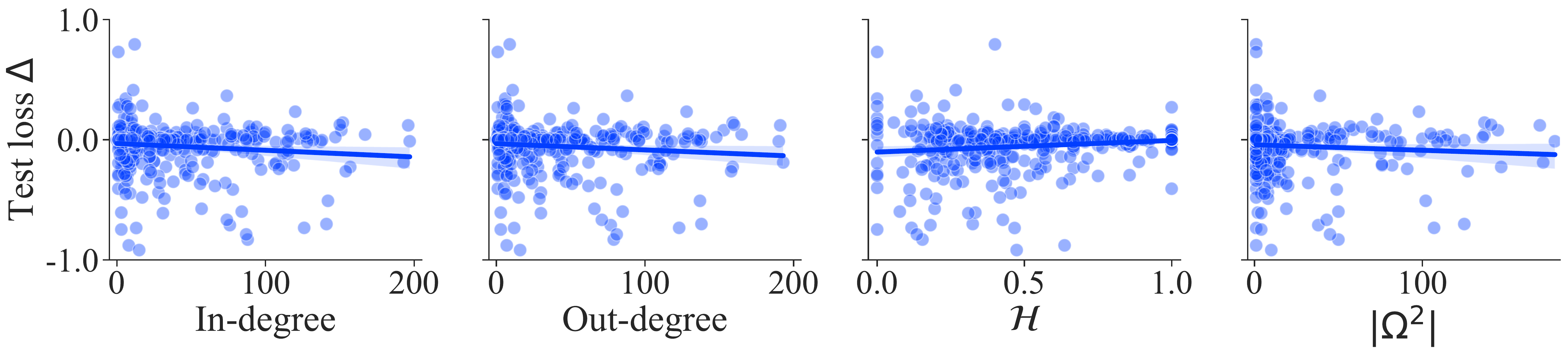}
    \caption{Air results.}
    \label{fig:ncfeatures}
\end{subfigure}
\par\bigskip
\begin{subfigure}{\linewidth}
    \centering
    \includegraphics[width=\linewidth]{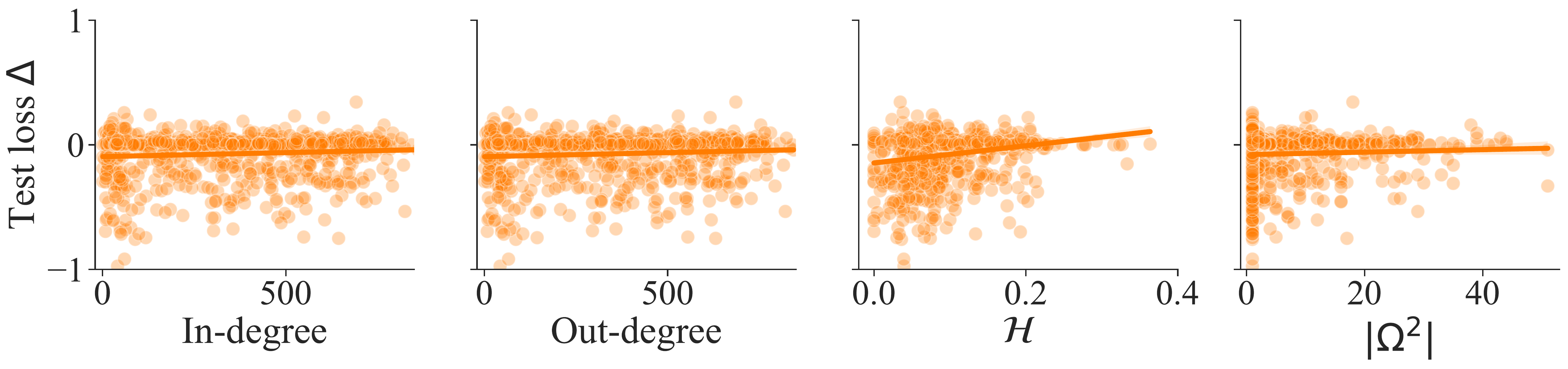}
    \caption{T2D results.}
    \label{fig:t2dfeatures}
\end{subfigure}
\par\bigskip
\begin{subfigure}{\linewidth}
    \centering
    \includegraphics[width=\linewidth]{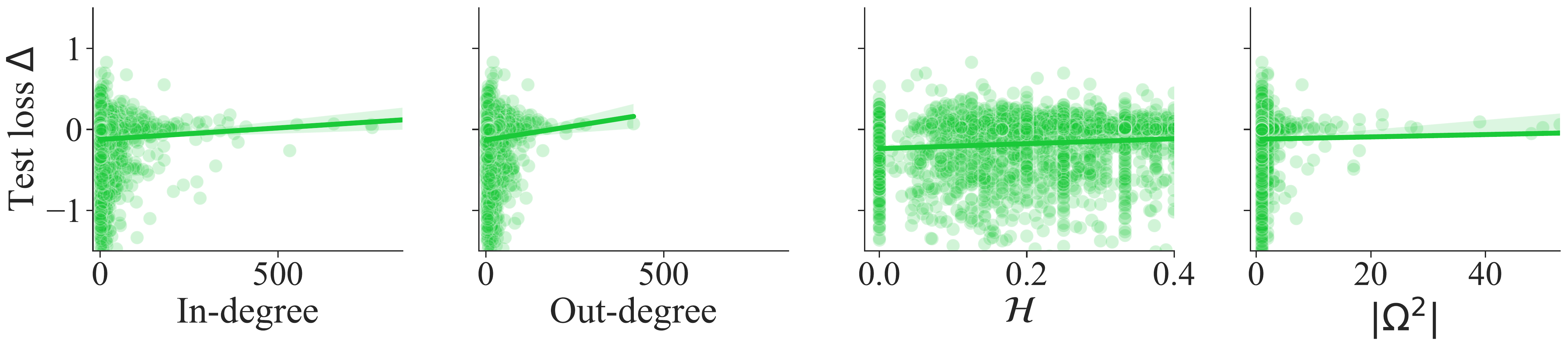}
    \caption{Wiki results.}
    \label{fig:wikifeatures}
\end{subfigure}
\par\bigskip
\begin{subfigure}{\linewidth}
    \centering
    \includegraphics[width=\linewidth]{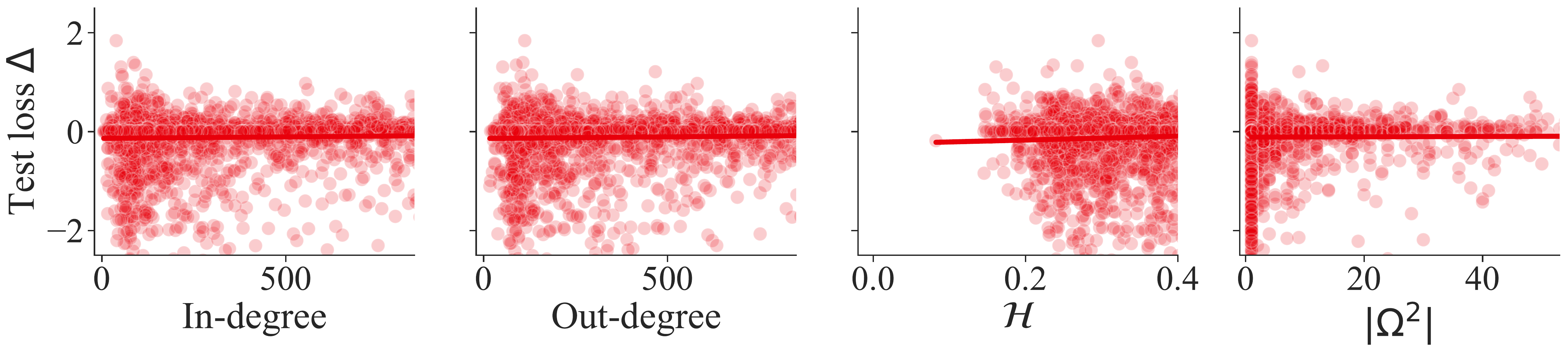}
    \caption{Mag results.}
    \label{fig:nyfeatures}
\end{subfigure}
\caption{Delta in test loss for \propc~compared to GraphSAGE (the base GNN) for node classification (i.e., values $< 0$ mean that GraphSAGE produced higher loss than \propc), plotted as a function of key node features in $\myg_1$. Out-degree and in-degree are unweighted. Each point represents a node in the test set. All five testing folds are included. Shading represents a 95\% CI for the regression line.}
\label{fig:features}
\end{figure}

To better understand \prop's performance, we analyzed the node classification predictions by plotting the change in test loss for each node (Figure \ref{fig:features}). We hypothesized that \prop would provide the strongest improvements on nodes with higher out-degree and larger higher-order families ($|\Omega^2|$); however, this was not consistently true for all data sets. Instead, we found that a node's homophily ($\mathcal{H}$, defined as the fraction of neighbors that have the same class) was the best indicator of performance improvement. This observation is noteworthy for future work, especially since most GNNs struggle to model graphs with low homophily \citep{zhu2020beyond}. 

\clearpage
\section{Additional considerations on HON construction} \label{sec:growhon}
\begin{figure}[ht]
    \centering
    \begin{subfigure}{.25\linewidth}
    \includegraphics[width=\linewidth]{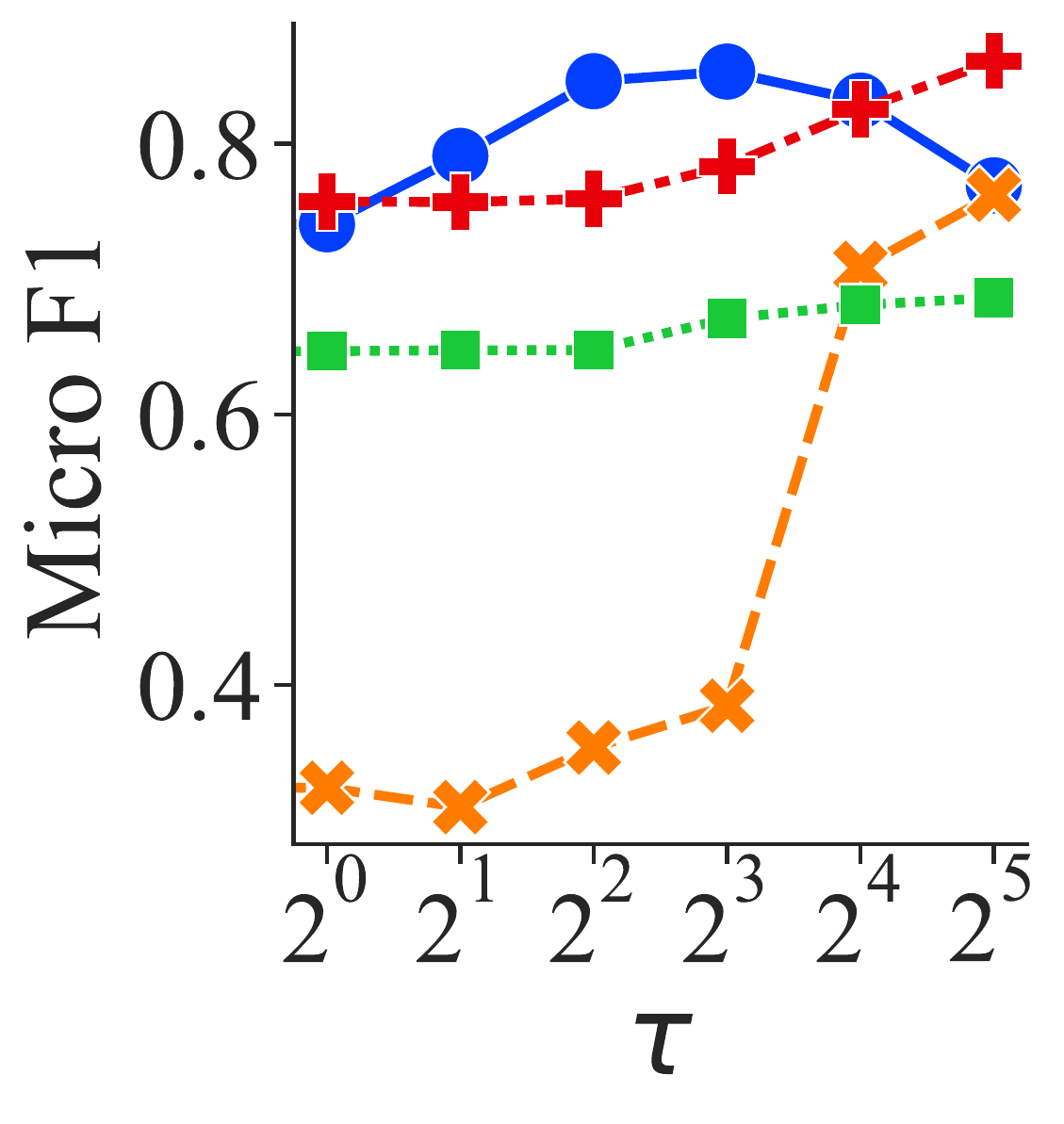}
    \caption{Node classification.}
    \end{subfigure}
    \begin{subfigure}{.25\linewidth}
    \includegraphics[width=\linewidth]{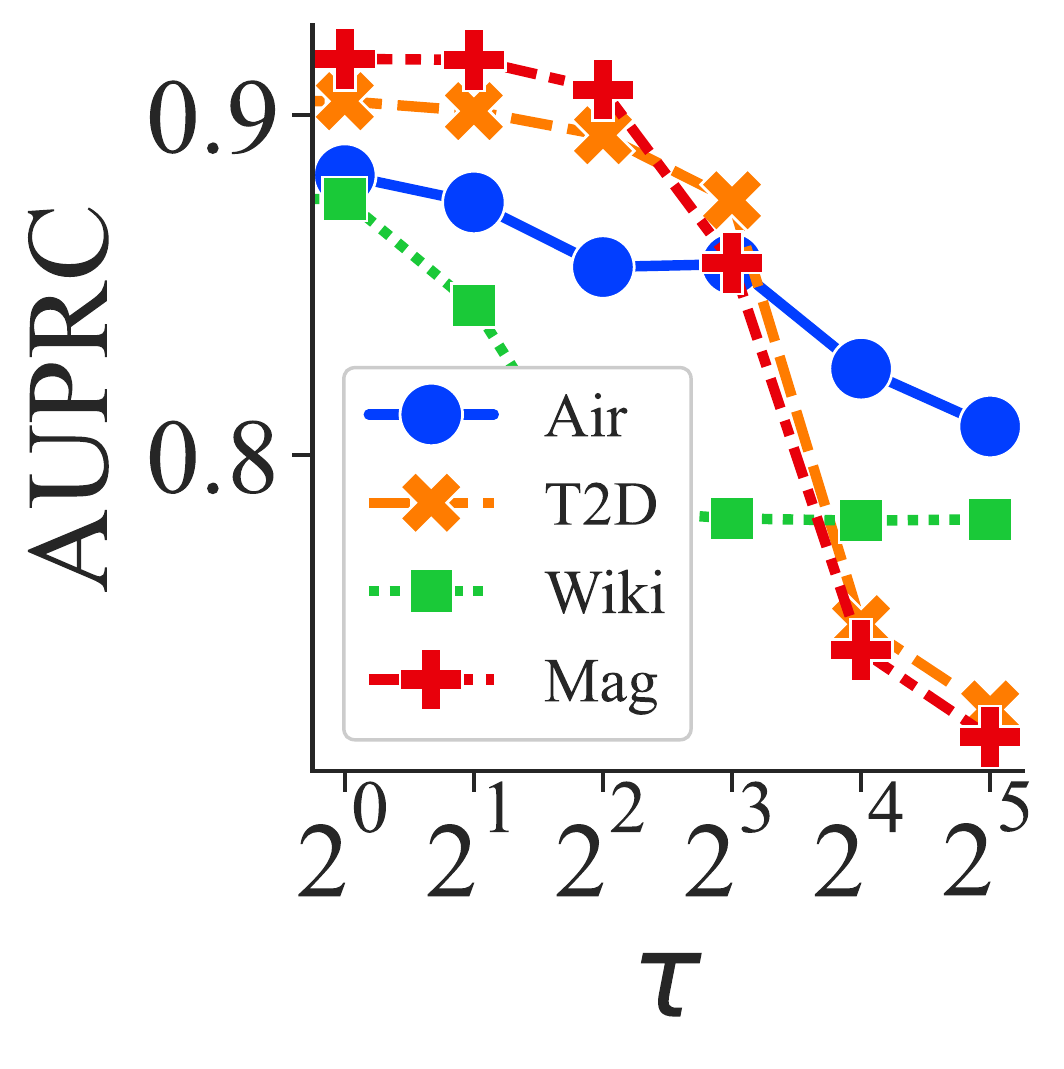}
    \caption{Link prediction.}
    \end{subfigure}
    \caption{Performance as a function of $\tau$. Each point represents the mean of all 5 testing folds.} 
    \label{fig:taus}
\end{figure}

We used GrowHON \citep{krieg2020growhon} to construct all HONs. GrowHON provides a parameter $\tau$, which helps regularize
the graph by raising the KL-divergence threshold required for a  new higher-order node to be created, and we found that the choice of $\tau$ impacted \prop's success in each task. Figure \ref{fig:taus} shows the results of varying $\tau$ for each data set and task. In general, the optimal $\tau$ value varied between data sets for classification tasks, while for link prediction, a lower $\tau$ generally yielded better results. We did not tune \prop's hyperparameters for these experiments, so it is likely that further tuning would reduce the difference between each model. However, these observations have important consequences for future work, especially with respect to integrating prior studies on graph structure learning, which have argued that different graph representations may be useful for different predictive tasks \citep{brugere2020inferring}. Ideally, we could incorporate HON construction into the learning framework so that the model could learn the structure that is best-suited to each task.

\end{document}